\documentclass[10pt,letterpaper]{article}

\usepackage{arxiv/techreport}
\usepackage{afterpage}
\usepackage{placeins}

\usepackage{multirow}
\usepackage{colortbl}
\usepackage{placeins}

\usepackage[normalem]{ulem}

\makeatletter
\renewcommand\paragraph{\@startsection{paragraph}{4}{\z@}{\z@ \@plus .5ex}{-1em}{\normalfont\normalsize\bfseries}}
\makeatother

\usepackage[pagebackref,breaklinks,colorlinks,allcolors=reportblue]{hyperref}
\hypersetup{
  pdftitle={Endogenous and Exogenous Memory for World Action Models},
  pdfauthor={Zerui Gong, Marina Barannikov, Zhonghua Wu, Shanshan Lu, Junjie Gao, Zhe Yang, Jinglin He, Quanzhou Li, Weijie Qiao, Qinyue Li, Qingyi Tao, Chen Change Loy}}

\newlength{\arxivcolw}
\newif\ifarxivnarrow
\AtBeginEnvironment{figure}{\arxivnarrowtrue}
\AtBeginEnvironment{table}{\arxivnarrowtrue}
\AtBeginEnvironment{figure*}{\arxivnarrowfalse}
\AtBeginEnvironment{table*}{\arxivnarrowfalse}
\makeatletter
\g@addto@macro\@floatboxreset{\ifarxivnarrow\setlength{\linewidth}{\arxivcolw}\fi}
\makeatother

\newcounter{arxivsupppage}
\newcommand{\maketitlesupplementary}{%
  \setcounter{page}{\value{arxivsupppage}}%
  \begin{center}
    {\fontsize{17}{21}\selectfont\bfseries\sffamily Supplementary Material\par}
  \end{center}
  \vspace{0.5em}%
  \suppressfloats[t]}

\title{Endogenous and Exogenous Memory for World Action Models}
\teamlogo{\includegraphics[height=1.05cm]{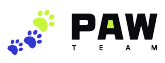}}
\author{%
  Zerui~Gong\aff{1*},
  Marina~Barannikov\aff{2*},
  Zhonghua~Wu\aff{1*\textdagger},
  Shanshan~Lu\aff{2},
  Junjie~Gao\aff{1},
  Zhe~Yang\aff{2},
  Jinglin~He\aff{1},
  Quanzhou~Li\aff{2},
  Weijie~Qiao\aff{1,2},
  Qinyue~Li\aff{1},
  Qingyi~Tao\aff{1},
  Chen~Change~Loy\aff{2,3}}
\affiliations{\aff{1}SenseTime Research, Singapore,\enspace
  \aff{2}S-Lab, Nanyang Technological University,\enspace
  \aff{3}ACE Robotics}
\authornotes{\aff{*}Equal contribution,\enspace \aff{\textdagger}Project lead}
\reportlinks{%
  \reportlink{Project Page}{https://pawteam-eai.github.io/em-wam/}}

\begin{document}

\begin{titlecard}
\begin{abstract}
World action models (WAMs) are effective at short-horizon manipulation skills, but extending them to long-horizon tasks requires remembering earlier information that is no longer visible.
Existing methods typically either retain all past frames, or compress frames by their position in the episode rather than by their content, or select what to keep by content but rely on additional supervision.
Moreover, even when the relevant frame is kept, an unexpected event, which cannot be predicted from the observed history, is often visible in a small region for only a few frames, so the policy may not attend to it.
To address both limitations, we present EM-WAM, a world action model with two complementary memories, which use the video expert's own signals without additional supervision.
Specifically, the \textbf{endogenous memory} is driven top-down by the task goal and records how the state changes as the task progresses. It uses a state-refresh score from the gated layers of the video expert to retain informative frames in full and compress the others into a few visual summary tokens.
The \textbf{exogenous memory} is driven bottom-up by unexpected events. It stores these events and uses them to modulate the action expert, even after they are no longer visible.
To evaluate how a policy maintains and updates its memory, we introduce RMBench-Dynamic, a benchmark of eight dual-arm manipulation tasks with unexpected events.
Experiments show that EM-WAM achieves state-of-the-art performance on both RMBench (81.7\%) and RMBench-Dynamic (58.8\%), and succeeds in 57 of 60 real-robot trials.
\end{abstract}

\end{titlecard}

\suppressfloats[t]
\begin{figure*}[t]
  \centering
  \includegraphics[width=\linewidth]{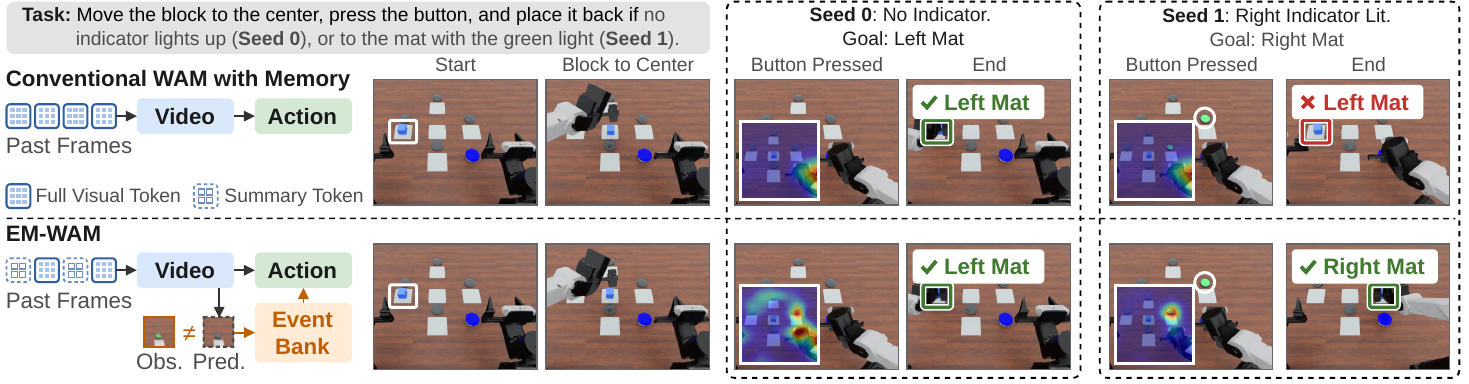}
  \caption{\textbf{Remembering a transient visual cue.} The block starts on the left mat. The robot moves it to the center and presses the button, and a brief visual cue then determines the destination. If no indicator lights up (Seed~0), the block returns to its original left mat. If the right indicator lights up (Seed~1), it goes to the right mat. A conventional WAM with memory (\eg, LingBot-VA~\cite{li2026lingbotva}) retains the past frames, but its action expert attends mainly to the robot's own arm and misses the cue, and it chooses the left mat in both cases. EM-WAM instead localizes the lit indicator from the discrepancy between its prediction and the observation, writes this region into the event bank, and uses it to modulate the action expert. As a result, it places the block on the correct mat, and its action expert attends to the lit indicator, as shown in the heatmaps.}
  \label{fig:teaser}
\end{figure*}

\section{Introduction}
\label{sec:intro}

World action models (WAMs) jointly predict future observations and actions, and they can now perform many short-horizon manipulation skills~\cite{li2025uva,zhu2025uwm,kairos2026kairos,li2026lingbotva}.
For long-horizon tasks, a high-level planner (System~2) is usually used to split the task into subtasks for a low-level policy (System~1) to execute~\cite{shi2025hirobot,sridhar2026memer,torne2026mem,han2025robocerebra,sensenova2026gpt6astra}.
However, System~2 runs at a much lower frequency than the robot controller, so a WAM that serves as System~1 executes each subtask over hundreds of control steps and needs a memory of its own for this execution.
For example, when System~2 asks the robot to add two spoons of salt to a soup, the robot must remember how many spoons it has added, since the salt dissolves and leaves no visible trace. If a person adds a spoon of salt in the meantime, the robot must update this count.

Existing methods for retaining visual information in WAMs often condition on long histories of uncompressed past observations~\cite{li2026lingbotva}.
As a result, GPU memory and computation grow linearly with the length of the episode.
To reduce this cost, some methods keep only the first and the most recent frames in full, and compress the other frames into a few latent tokens~\cite{yang2026memorywam}.
These methods decide which frames to keep in full by their position in the episode, although the actual importance of a frame depends on what it shows.
Other memory-based policies select what to write into memory, such as keyframes or language summaries, but rely on additional supervision or a separate model to make this selection~\cite{sridhar2026memer,torne2026mem}.
A memory should therefore decide from its content, without additional supervision, whether to keep a frame in full, keeping the informative frames and compressing the redundant ones.

Moreover, keeping informative frames is still not enough for unexpected events, which cannot be predicted from the observed history, such as an object moved by a person, a revised goal, or an action with no effect.
Later actions depend on such an event, but it often covers a small region for a few frames, or a small fraction of a long observation history, so keeping the frame does not ensure that the policy attends to the change. For example, in \cref{fig:teaser}, the robot must place a block on the mat next to the indicator that lights up while a button is pressed.
A conventional WAM (\eg, LingBot-VA~\cite{li2026lingbotva}) retains every past frame in full, yet its attention stays on its own arm while the indicator is lit, and it fails when the right indicator lights up.
This observation suggests that its attention does not reach the scene change, even though the frame is kept.

Prior work detects such changes with failure monitors~\cite{guo2023doremi,zhou2025codeasmonitor} or re-plans with a language model in a hierarchical system~\cite{huang2023innermonologue,liu2023reflect,shi2025hirobot}.
Recent WAMs also compare their predicted future with the actual observation to decide when to re-plan~\cite{wang2026whentotrust,xue2026hitacwam}, or rank candidate futures to decide which one to execute~\cite{zhang2026wcd}.
Yet the same discrepancy between the predicted and the actual observation is large where such a change occurs, so it can indicate, without additional supervision, what the policy should remember.

\begin{figure*}[t]
  \centering
  \includegraphics[width=\linewidth]{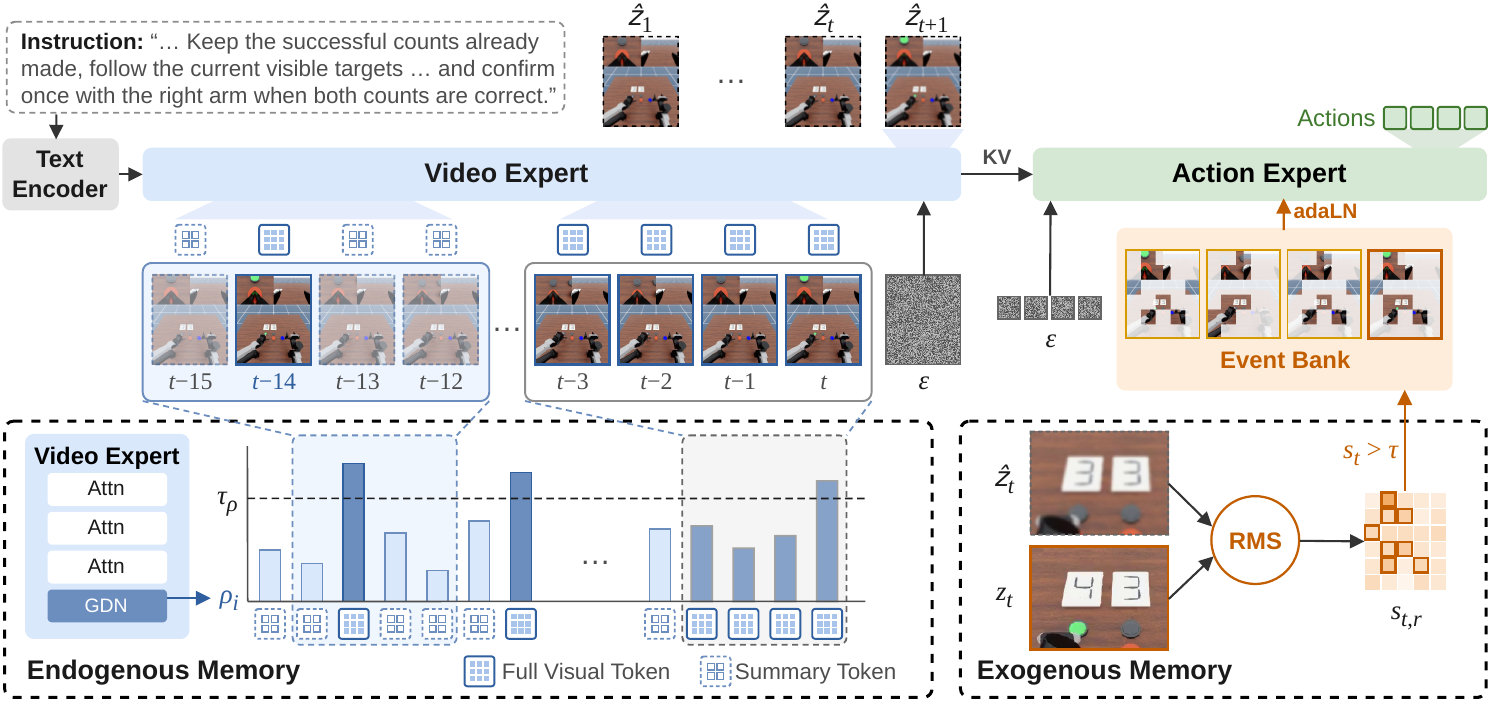}
  \caption{\textbf{EM-WAM overview.} The video expert interleaves attention layers with gated layers (GDN), and the action expert attends to its keys and values (KV). Endogenous memory: when an observation leaves the recent window, it is retained in full if its state-refresh score from the last gated layer reaches a threshold, and only its summary tokens are kept otherwise. Exogenous memory: when the discrepancy between an observation and its prediction from the previous step exceeds a threshold, the regions with the largest discrepancy are written to an event bank, which the action expert reads by cross-attention and uses to modulate its layers through adaptive layer normalization (adaLN). }
  \label{fig:method}
\end{figure*}

Based on these observations, we propose \textbf{EM-WAM}, a world action model that augments its video expert and action expert with two memory mechanisms, both using signals that the video expert already produces~(\cref{fig:method}).
Specifically, the \textbf{endogenous memory} is driven top-down by the task goal and records how the state changes as the task progresses. It uses a state-refresh score from the gated layers of the video expert to detect informative frames, retaining them in full. The remaining frames are compressed into a few \emph{visual summary tokens}.
The \textbf{exogenous memory} is driven bottom-up by unexpected events. It stores these events in an event bank, which modulates the action expert even after the event is no longer visible.
In \cref{fig:teaser}, the lit indicator is such an event; once it is stored, the attention of the action expert reaches this change, and the robot places the block on the correct mat.

In existing memory benchmarks, task-relevant information mostly stays valid once observed.
To evaluate whether a policy can handle unexpected events, we introduce \textbf{RMBench-Dynamic}, a benchmark of eight dual-arm manipulation tasks~(\cref{fig:rmbenchpp_tasks}).
Each task is built around an unexpected event, such as a revised goal, a moved object, an action with no effect, or brief feedback from an exploratory action.
To complete a task, the model must keep the past information that is still valid and update its information about the current scene, so that it generates the correct actions.

\enlargethispage{\baselineskip}
In summary, our \textbf{main contribution} is EM-WAM, a world action model with two memories: an endogenous memory that keeps informative past frames in full and compresses the rest, reducing visual tokens by 38.8\% compared with keeping the full history on RMBench-Dynamic%
, and an exogenous memory that records unexpected events and modulates the generation of actions, improving the average success rate on RMBench-Dynamic by 8.0 points over the model without the exogenous memory%
. We also introduce RMBench-Dynamic, a benchmark of eight dual-arm manipulation tasks that evaluates whether a policy can handle unexpected events. 
EM-WAM achieves state-of-the-art performance on RMBench-Dynamic (58.8\%, 14.2 points above the previous state-of-the-art method, LingBot-VA~\cite{li2026lingbotva}) and RMBench~\cite{chen2026rmbench} (81.7\%).
We also validate our method on a real dual-arm robot, where it succeeds in 57 of 60 trials over two tasks, compared with 27 of 60 for LingBot-VA.

\section{Related Work}
\label{sec:related_work}

\noindent\textbf{World Action Models.}
Video-based policies plan with generated videos~\cite{du2023unipi}, pretrain on future-frame prediction~\cite{wu2024gr1,cheang2024gr2}, or condition actions on predicted visual states~\cite{tian2025seer,hu2025vpp}.
World action models (WAMs) couple video and action generation in one model~\cite{li2025uva,zhu2025uwm,kairos2026kairos}; some skip video generation at test time to reduce latency~\cite{yuan2026fastwam}.
For long horizons, LingBot-VA~\cite{li2026lingbotva} retains every past observation in full,
MemoryWAM~\cite{yang2026memorywam} keeps the first and the most recent observations in full and compresses the others into gist tokens, HiMem-WAM~\cite{sun2026himemwam} learns when to write memory from skill-boundary pseudo-labels, and RM-WAM~\cite{miao2026rmwam} learns how much memory to inject.
These methods decide what to keep by position or with additional supervision.
Other WAMs compare the predicted future with the actual observation at test time to decide when to re-plan~\cite{wang2026whentotrust,xue2026hitacwam} or which sampled future to execute~\cite{zhang2026wcd}.
EM-WAM builds its video expert on Kairos~\cite{kairos2026kairos} and adds an endogenous and an exogenous memory, both driven by signals from this expert, without additional supervision.

\noindent\textbf{History Compression and Selective State.}
Conditioning imitation policies on long observation histories is difficult~\cite{wen2020copycat,wen2021keyframe,torne2025pasttoken}.
Memory-based policies retain a window of recent observations~\cite{fang2025sam2act,torne2026mem}, write to a separate memory such as memory banks, keyframes, language summaries or a revisable task state~\cite{shi2026memoryvla,sridhar2026memer,yang2026eventvla,torne2026mem,senthilkumaran2026remm}, or carry a recurrent state through the episode~\cite{li2026rememvla,chung2026liberomem,zhao2026dram}.
Language models bound their caches by eviction or compression~\cite{xiao2024streamingllm,zhang2023h2o,rae2020compressive}, and selective state-space models and linear attention with delta-rule or gated updates let the current input control what the stored state keeps~\cite{gu2024mamba,dao2024mamba2,schlag2021linear,yang2024deltanet,yang2025gateddeltanet}.
In EM-WAM, the endogenous memory reads this control from the gated layers of the video expert to decide which observations to keep in full, and compresses the others into summary tokens~\cite{yang2026memorywam}.

\noindent\textbf{Unexpected Changes and Prediction Discrepancy.}
Beyond WAMs, robots detect unexpected changes with failure monitors~\cite{guo2023doremi,zhou2025codeasmonitor} or re-plan with a language model in a hierarchical system~\cite{huang2023innermonologue,liu2023reflect,shi2025hirobot}. In perception and sequence models, the discrepancy between predictions and observations has been used to segment events and to form memories~\cite{aakur2019perceptual,behrouz2025titans,fountas2025emllm}. In EM-WAM, the exogenous memory uses this discrepancy to select the regions it stores in an event bank, which modulates the action expert after the event is no longer visible.

\noindent\textbf{Robot Manipulation Benchmarks.}
Simulated benchmarks cover multi-task learning~\cite{yu2020metaworld,james2020rlbench}, language-conditioned control~\cite{mees2022calvin,liu2023libero}, dual-arm manipulation~\cite{chen2026robotwin2}, robustness to visual variations~\cite{fei2026liberoplus}, and joint simulated and real-world evaluation~\cite{chen2026robodojo}. Memory benchmarks, including RMBench~\cite{chen2026rmbench} and others~\cite{fang2025sam2act,cherepanov2026mikasarobo,chung2026liberomem,dai2026robomme,sun2026memobench}, make success depend on information that is no longer observable when it is needed.
In most of these benchmarks, task-relevant information remains valid once observed.
RoboCerebra~\cite{han2025robocerebra} and RoboMME~\cite{dai2026robomme} introduce mid-episode scene changes, but RoboCerebra mainly evaluates high-level planners, and the changes in RoboMME follow a small set of systematic patterns.
RMBench-Dynamic builds every task around unpredictable in-episode events, requiring an end-to-end policy to both retain valid past information and update what has become outdated.

\section{Method}
\label{sec:method}

EM-WAM equips a world action model with two new memories~(\cref{fig:method}).
The endogenous memory (\cref{sec:method_endogenous}) uses a state-refresh score to keep in full the informative observations and compress the others into a few summary tokens.
The exogenous memory (\cref{sec:method_exogenous}) stores the regions with a large discrepancy %
 between the prediction of the video expert and the observation in an event bank, which modulates the action expert.

%-------------------------------------------------------------------------------
\subsection{Endogenous Memory}
\label{sec:method_endogenous}

Our video expert builds on the Kairos backbone~\cite{kairos2026kairos}, where every fourth layer is a gated linear-attention layer of the GatedDeltaNet type~\cite{yang2025gateddeltanet} and the other layers are attention layers.
Both experts condition on the history of observations through the attention layers. Retaining every observation in full makes the cost grow with the length of the episode, so the endogenous memory retains in full the recent observations and the informative earlier ones.
The endogenous memory keeps the history in a hybrid token cache $\mathcal{C}_t$ read by the attention layers of the video expert, where an observation is stored either as its full visual tokens or as a few summary tokens. A \emph{state-refresh score} $\rho_t$ decides between the two. The score is read from the decay gates of the gated layers, so the selection needs no additional supervision.

\noindent\textbf{Hybrid Token Cache.}
Each observation is encoded by the VAE into a latent $z_t$ and has two representations, full visual tokens and summary tokens.
We follow MemoryWAM~\cite{yang2026memorywam} to obtain the summary tokens.
The latent $z_t$ is patchified into $L$ full visual tokens, followed by $M \ll L$ learnable summary tokens ($M = 4$ and $L = 120$ in our experiments).
The full visual tokens are these VAE features without further compression.
The summary tokens attend to the visual tokens of their own observation and to the cache at that step, so they form a content-dependent compression of the observation.
Every token keeps an absolute position in the 3D rotary embedding, so cached entries are never re-encoded when the visual tokens of an earlier observation are removed; summary tokens take the temporal position of their observation and a fixed spatial marker.
Unlike MemoryWAM~\cite{yang2026memorywam}, we choose which representation an observation keeps from its content, depending on whether it is informative.

\noindent\textbf{State-Refresh Score.}
Each gated layer compresses the whole history into a recurrent state $S$ per head, and its decay gate indicates whether a token brings new content.
Before writing each token $x_j$, the layer multiplies $S$ by this gate.
The decay gate is $\alpha_j = \exp(-e^{A} \Delta_j) \in (0, 1)$, controlled by the step size $\Delta_j = \operatorname{softplus}(w_a^{\top} x_j + b_\Delta)$, where $w_a$, $b_\Delta$ and $A$ are learned parameters~\cite{gu2024mamba,dao2024mamba2}.
The step size depends on the input, so the layer learns a large $\Delta_j$, which drives $\alpha_j$ toward zero and lets $S$ decay, when $x_j$ brings content that $S$ does not hold.

The state-refresh score aggregates these decays over the visual tokens of an observation.
The decay gates of an observation's tokens multiply within each head, so the sum of their step sizes in that head sets how much of its state carried from earlier observations survives.
The larger this sum, the more of the earlier state the observation replaces, and the more new information it brings.
We compute the score $\rho_t$ of observation $t$ at the last gated layer when the observation is encoded:
\begin{equation}
  \rho_t = \Bigl\lVert \frac{1}{L} \sum_{j \in \mathcal{V}_t} \boldsymbol{\Delta}_j \Bigr\rVert_2 ,
  \label{eq:delta_score}
\end{equation}
where $\mathcal{V}_t$ indexes the $L$ visual tokens of observation $t$ and $\boldsymbol{\Delta}_j$ stacks the step sizes of all heads.
Since $A$ is fixed per head, a larger mean step size means that more of the head's state is refreshed, and the norm combines all heads into one score.

\noindent\textbf{Retention Decision.}
The $R$ most recent observations stay in full, since predicting the next observation and the next actions requires the precise current state of the scene, such as the pose of the arms and the positions of objects. When observation $i$ leaves this window, it is kept in full if $\rho_i \ge \tau_\rho$ for a fixed threshold $\tau_\rho$; otherwise its visual tokens are discarded and only its $M$ summary tokens remain.
The cache at step $t$ is therefore
\begin{equation}
  \mathcal{C}_t = \left\{
  \begin{array}{@{}l@{\quad}l@{}}
    Z_i, & t-R < i \le t, \\
    Z_i, & i \le t-R,\ \rho_i \ge \tau_\rho, \\
    G_i, & i \le t-R,\ \rho_i < \tau_\rho
  \end{array}
  \right\}_{i \le t},
  \label{eq:context}
\end{equation}
where $Z_i$ and $G_i$ denote the full visual tokens and summary tokens of observation $i$.
The observations retained in full are those right after a subtask is completed~(\cref{fig:memory_vis}a).
The decision is made once per observation and depends only on that observation and its past, so it is made online at inference and fixed before each training pass (\cref{sec:method_training}).

%-------------------------------------------------------------------------------
\subsection{Exogenous Memory}
\label{sec:method_exogenous}

The endogenous memory decides which observations to retain, but it does not ensure that the policy uses an unexpected event.
Such an event is often visible in a small region for a few observations. Even when the endogenous memory retains such an observation in full, the changed tokens are a small part of a long cache, and the attention of the action expert can miss them.
Since this event cannot be predicted from the history, it appears as a discrepancy between the prediction of the video expert and the observation.
The exogenous memory therefore writes the regions with the largest discrepancy into a separate event bank of $N_m$ slots, which is reset at the start of each episode, and uses the bank to modulate the action expert directly.

\noindent\textbf{Discrepancy as Surprise.}
At step $t-1$, the video expert predicts the next latent $\hat{z}_{t}$ from $\mathcal{C}_{t-1}$ with $K$ denoising steps.
Once $z_{t}$ is observed, we measure the discrepancy by the surprise $s_{t}$, the root mean square (RMS) of the prediction error,
\begin{equation}
  s_{t} = \operatorname{RMS}\bigl( \hat{z}_{t} - z_{t} \bigr),
  \label{eq:surprise}
\end{equation}
computed over all latent coordinates.
The surprise of an observation does not show where the change occurs. We therefore also compute the surprise of local regions.
The $L$ visual tokens of an observation (\cref{sec:method_endogenous}) are arranged on a spatial grid, and we group every $2 \times 2$ neighboring tokens into one region, which gives $N_r$ regions per observation.
The regional surprise $s_{t,r}$ is the RMS of the prediction error within region $r$.
The observation-level surprise decides whether an observation is written into the bank, and the regional surprise selects which of its regions are stored~(\cref{fig:memory_vis}b).

\noindent\textbf{Write Decision.}
The bank has only $N_m$ slots, while most observations follow the history and differ from the prediction only by small errors. The write decision therefore reserves the bank for the most surprising observations.
An observation is written only if its surprise exceeds a threshold $\tau$, the $q$-th quantile of a running set of past surprise scores. The quantile adapts the threshold to the scale of the prediction error, which varies across tasks and scenes. The threshold is fixed within an episode, updated between episodes, and initialized at deployment from a calibration rollout. 
A written observation takes a free slot. When the bank is full, it replaces the oldest slot as a whole, so the bank keeps the $N_m$ most recently written observations of the episode.

\noindent\textbf{Event Tokens.}
Each written observation is stored as $P$ event tokens, one for each of its most surprising regions. Storing only these regions focuses the memory on where the scene has changed and keeps the bank compact.
Each region is represented by the mean of the video expert's last-layer output tokens over the region, computed when the observed $z_{t}$ is encoded. This feature $m$ is projected and combined with an embedding of its metadata into an event token,
\begin{equation}
  e = \operatorname{LN}\bigl( W_c\, m + \phi( s_{t,r},\, s_{t},\, \text{age},\, \text{row},\, \text{col} ) \bigr),
  \label{eq:memory_token}
\end{equation}
where $W_c$ is a linear projection, $\phi$ embeds the metadata, $\operatorname{LN}$ is layer normalization, $s_{t,r}$ and $s_{t}$ are the regional and observation-level surprise, the age is the number of steps since the write (normalized by $N_m$), and $(\text{row}, \text{col})$ is the normalized location of the region in the frame.
The metadata tells the action expert how surprising the change was, how long ago it occurred, and where it is in the frame.

\begin{samepage}
\noindent\textbf{Action Expert Modulation.}
Rather than mixing the event tokens into the long history, where they could again be missed, we use the bank to modulate the action expert directly.
Let $H_t$ be the input embeddings of the action tokens at step $t$ and $E_t$ the event tokens.
The action tokens read the bank by cross-attention~\cite{dai2026robomme}:
\begin{equation}
  r_t = \operatorname{Attn}\bigl( H_t W_Q,\ [\, n ; E_t \,] W_K,\ [\, n ; E_t \,] W_V \bigr).
  \label{eq:read}
\end{equation}
\end{samepage}
Here, $W_Q$, $W_K$ and $W_V$ are learned projections, and $n$ is a learned null token that the read can attend to when the bank is empty or irrelevant.
The result $r_t$ passes through a zero-initialized linear layer and is added to the timestep modulation shared by all blocks of the action expert, which sets the shift, scale and gate of their adaptive layer normalization (adaLN)~\cite{peebles2023dit}.

%-------------------------------------------------------------------------------
\subsection{Training Strategy}
\label{sec:method_training}

We train EM-WAM in two stages, one for each memory. The exogenous memory relies on the predictions of the video expert, so Stage~1 first trains both experts with the endogenous memory, and Stage~2 then adds the exogenous memory.

\noindent\textbf{Stage 1: Endogenous Memory.}
We initialize the video expert from the pretrained Kairos~\cite{kairos2026kairos}.
Each training iteration runs two passes over the episode.
(1)~A gradient-free pass computes the state-refresh score $\rho_i$ of every observation and fixes its retention decision, which determines the cache $\mathcal{C}_t$ of each step.
(2)~The training pass processes the whole episode at once, with the attention of each step masked to its cache $\mathcal{C}_t$, so that each step sees the same cache as at inference.
It trains both experts with a video loss and an action loss,
\begin{equation}
  \mathcal{L}_{1} = \textstyle\sum_{t} \bigl( \ell( z_{t+1} \mid \mathcal{C}_t ) + \ell( a_t \mid \mathcal{C}_t ) \bigr),
  \label{eq:losses}
\end{equation}
where $\ell$ is the flow-matching loss~\cite{lipman2023flowmatching,liu2023rectifiedflow}, the ground truth of the video loss is the VAE latent $z_{t+1}$ of the next observation, and the ground truth of the action loss is the action chunk $a_t$ of the demonstration.

\noindent\textbf{Stage 2: Exogenous Memory.}
Stage~2 starts from the Stage-1 weights and adds the event bank.
Its gradient-free pass additionally predicts each next latent with $K$ denoising steps and computes the surprise, which selects the written observations and their regions (\cref{sec:method_exogenous}) and thus determines the event tokens $E_t$ of each step.
The training pass then optimizes
\begin{equation}
  \mathcal{L}_{2} = \textstyle\sum_{t} \bigl( \ell( z_{t+1} \mid \mathcal{C}_t ) + \ell( a_t \mid \mathcal{C}_t, E_t ) \bigr).
  \label{eq:losses_stage2}
\end{equation}

\section{RMBench-Dynamic}
\label{sec:rmbenchpp}

RMBench-Dynamic extends the memory evaluation of RMBench~\cite{chen2026rmbench} to unpredictable in-episode events, such as temporary signals in the environment or a perturbed scene. Such an event cannot be predicted from the history, so the policy must notice it and update its memory.
Every task is built around such an event, and the manipulation is kept simple so that the tasks stress memory.
Five of the RMBench-Dynamic tasks extend RMBench tasks with such an event; the other three are new.

\begin{figure*}[!htbp]
  \centering
  \includegraphics[width=\linewidth]{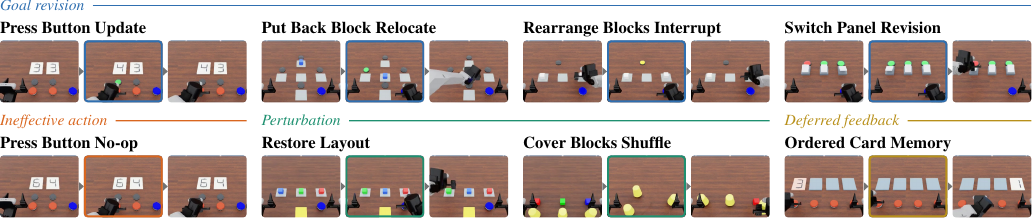}
  \caption{\textbf{The eight tasks of RMBench-Dynamic.}
  For each task, we show the cue, the moment it changes (outlined in the color of its event type), and a later moment when the policy must act on it from memory.
  Tasks are grouped by the type of in-episode event (\cref{sec:rmbenchpp_tasks}); Sec.~A of the supplement gives the details of each task.}
  \label{fig:rmbenchpp_tasks}
\end{figure*}

\label{sec:rmbenchpp_tasks}

\noindent\textbf{Event Types.}
We group the tasks by their primary event type (\cref{fig:rmbenchpp_tasks}).
In \emph{goal revision} (Press Button Update, Put Back Block Relocate, Rearrange Blocks Interrupt, Switch Panel Revision), the target changes during execution, so the policy must revise previously observed information.
In \emph{ineffective action} (Press Button No-op), an action sometimes has no effect, and the policy must notice this and repeat it.
In \emph{perturbation} (Restore Layout, Cover Blocks Shuffle), the environment alters the scene after the policy has observed it.
In \emph{deferred feedback} (Ordered Card Memory), the result of an exploratory action must be taken into account even after the visual evidence disappears.
In several tasks the event is conditional, \eg, an indicator may or may not redirect the goal, so the policy has to watch for the cue.
All cues and feedback are visual, \eg, number cards, indicators and block colors, and most are shown only briefly.
Per-task rules and success criteria are in Sec.~A.3 of the supplementary material.

\FloatBarrier
\noindent\textbf{Execution and Data.}
\label{sec:rmbenchpp_protocol}
RMBench-Dynamic runs in Isaac Sim with two ARX X5 arms, observed by a head camera and one camera on each wrist.
RMBench-Dynamic executes actions and renders observations at the same fixed rate during both data collection and evaluation, so a brief signal spans the same number of observations in both.
Each task comes with 50 demonstrations recorded at 25\,Hz.
Evaluation uses 100 held-out seeds per task, and the event schedule of every seed is fixed in advance.
Details of the executor are in Sec.~A.1 of the supplementary material, and per-task statistics are in Tab.~A.

\providecommand{\realTaskEndo}{Cover Blocks}
\providecommand{\realTaskExo}{Restore Layout}

\section{Experiments}
\label{sec:experiments}

%-------------------------------------------------------------------------------

\subsection{Experimental Setup}
\label{sec:exp_setup}

\noindent\textbf{Benchmarks.}
We evaluate EM-WAM on RMBench~\cite{chen2026rmbench} and RMBench-Dynamic~(\cref{sec:rmbenchpp}).
RMBench has nine dual-arm manipulation tasks whose success depends on the information observed earlier in the episode.
On both benchmarks, a separate model is trained per task.
We report the success rate over 100 evaluation episodes per task, with seeds 100000--100099 on RMBench and 100001--100100 on RMBench-Dynamic.

\noindent\textbf{Implementation Details.}
We use the pretrained Kairos~\cite{kairos2026kairos} as the backbone, including a 4B video DiT, a Qwen2.5-VL-7B text encoder~\cite{bai2025qwen25vl}, and a Wan2.1 video VAE~\cite{wan2025wan}.
The video expert has 32 layers of width 2560, where every fourth layer is a gated linear-attention layer and the others attend to the whole cache $\mathcal{C}_t$ (\cref{eq:context}).
The action expert shares the same architecture as the video expert but uses a reduced hidden dimension of 640 and full attention in every layer, and it is initialized from the video expert.
The input observation is a single mosaic image of the three cameras: the left and right wrist views ($64 \times 80$ each) are placed side by side above the head view ($128 \times 160$), giving a $192 \times 160$ image.
Both arms are controlled in absolute joint space.
We set the action horizon to $h=16$.
Each chunk spans 16 frames, four of which (every fourth) are encoded into a single latent frame, so each step $t$ of the memories corresponds to one chunk.
All hyperparameters are listed in the supplement (Sec.~C).

%-------------------------------------------------------------------------------
\subsection{Comparison with State-of-the-Art Methods}
\label{sec:exp_main}

\noindent\textbf{RMBench-Dynamic.}
\cref{tab:rmbenchpp} reports the per-task success rates on RMBench-Dynamic.
EM-WAM reaches the highest average success rate of 58.8\%, 14.2 points above the strongest baseline, LingBot-VA~\cite{li2026lingbotva} (44.6\%).
The endogenous memory alone (EM-WAM w/o exogenous) compresses the observations with a low state-refresh score into summary tokens, yet it reaches 50.8\%, 6.2 points above LingBot-VA, which retains every past observation in full.
Its gain over LingBot-VA is largest on the three tasks where more than half of the visual tokens are summarized: selecting what to keep in full helps most when few observations are decisive.
Adding the exogenous memory improves all eight tasks. The gain is large when an event leaves no visible trace but a later action depends on it, as in \textit{Restore Layout} (54\% to 81\%) and \textit{Put Back Block Relocate} (86\% to 98\%). It is small when the change stays in view or the action depends on a count or an order built over the episode.
Policies without long-horizon memory ($\pi_{0.5}$~\cite{black2025pi05} and Fast-WAM~\cite{yuan2026fastwam}) and the generalist VLA DM0.5~\cite{dexmal2026dm05} fail on most tasks that require tracking state updates.

\begin{table*}[!htbp]
  \centering
  \caption{\textbf{Results on RMBench-Dynamic.}
  Success rate (\%) per task.
  Best results are in \textbf{bold}.
  The last two rows give, for EM-WAM, the average numbers of summarized and full chunks per episode, and the reduction of visual tokens relative to keeping every chunk in full (higher is better).}
  \label{tab:rmbenchpp}
  \setlength{\tabcolsep}{3.6pt}
  \footnotesize
  \begin{tabular}{lccccccccc}
    \toprule
    Method
      & \shortstack{Press\\Button\\Update}
      & \shortstack{Put Back\\Block\\Relocate}
      & \shortstack{Rearrange\\Blocks\\Interrupt}
      & \shortstack{Switch\\Panel\\Revision}
      & \shortstack{Press\\Button\\No-op}
      & \shortstack{\vphantom{X}\\Restore\\Layout}
      & \shortstack{Cover\\Blocks\\Shuffle}
      & \shortstack{Ordered\\Card\\Memory}
      & Avg. \\
    \midrule
    $\pi_{0.5}$~\cite{black2025pi05}       & 1& 47& 31& 1& 3& 26& 0& 49& 19.8\\
    Fast-WAM~\cite{yuan2026fastwam} & 5& 32& 55& 17& 4& 9& 0& 49& 21.4\\
    DM0.5~\cite{dexmal2026dm05} & 2 & 40 & \textbf{66} & 3 & 1 & 26 & 16 & 52 & 25.8 \\
    LingBot-VA~\cite{li2026lingbotva} & 40 & 71 & 17 & 37 & 14 & 48 & 54 & \textbf{76} & 44.6 \\
    \midrule
    EM-WAM w/o exogenous           & 48 & 86 & 46 & 41 & 13 & 54 & 66 & 52 & 50.8 \\
    \rowcolor{gray!15}
    EM-WAM                             & \textbf{49} & \textbf{ 98} & 53 & \textbf{50} & \textbf{15} & \textbf{81} & \textbf{70} & 54 & \textbf{58.8} \\
    \quad\textit{Summarized / full chunks} & 9.1/22.9 & 16.2/10.2 & 14.5/11.5 & 9.2/66.6 & 5.2/26.9 & 12.9/22.1 & 39.6/25.4 & 17.4/17.3 & 15.5/25.4 \\
    \quad\textit{Token reduction (\%) $\uparrow$} & 28.1 & 59.2 & 53.9 & 10.1 & 16.1 & 35.6 & 58.9 & 48.6 & 38.8 \\
    \bottomrule
  \end{tabular}
\end{table*}

\begin{table}[!htbp]
  \centering
  \caption{\textbf{Results on RMBench.}
  Average success rate (\%) over the five $\mathrm{M}(1)$ tasks, the four $\mathrm{M}(n)$ tasks, and all nine tasks; tasks are grouped by memory complexity as in RMBench~\cite{chen2026rmbench}.
  Per-task results are in Tab.~B of the supplementary material.
  Results of the other methods are taken from MemoryWAM~\cite{yang2026memorywam}.
  Best results are in \textbf{bold}.}
  \label{tab:rmbench}
  \footnotesize
  \begin{tabular}{lccc}
    \toprule
    Method & $\mathrm{M}(1)$ & $\mathrm{M}(n)$ & Avg. \\
    \midrule
    $\pi_{0.5}$~\cite{black2025pi05}     & 14.4 & 5.5 & 10.4 \\
    Fast-WAM~\cite{yuan2026fastwam} & 1.4 & 11.5 & 5.9 \\
    LingBot-VA~\cite{li2026lingbotva} & 78.0 & 76.0 & 77.1 \\
    MemoryWAM~\cite{yang2026memorywam}   & 79.8 & 81.5 & 80.6 \\
    \midrule
    EM-WAM w/o exogenous       & 78.6 & \textbf{83.8} & 80.9 \\
    \rowcolor{gray!15}
    EM-WAM                               & \textbf{80.0} & \textbf{83.8} & \textbf{81.7} \\
    \bottomrule
  \end{tabular}
\end{table}

\noindent\textbf{RMBench.}
Following RMBench, \cref{tab:rmbench} reports the average success rates by memory-complexity group, and Tab.~B in the supplement lists the per-task results.
Policies without long-horizon memory, such as $\pi_{0.5}$~\cite{black2025pi05} and Fast-WAM~\cite{yuan2026fastwam}, fail on most tasks.
Among WAMs with long-horizon memory, EM-WAM w/o exogenous already outperforms LingBot-VA~\cite{li2026lingbotva} and MemoryWAM~\cite{yang2026memorywam}.
The gain is largest on tasks whose key cues appear at variable moments, such as \textit{Press Button} (91\% vs.\ 87\% for MemoryWAM and 84\% for LingBot-VA), which shows the benefit of selecting observations by an internal signal from the model rather than by their position.
Full EM-WAM reaches the highest average success rate of 81.7\%.

%-------------------------------------------------------------------------------

%-------------------------------------------------------------------------------
\subsection{Ablations}
\label{sec:exp_ablation}

We ablate the endogenous memory (\cref{tab:endogenous}) and the exogenous memory (\cref{tab:exogenous}) separately.

\label{sec:exp_endogenous}
\definecolor{ablK}{HTML}{C9F0E1}
\definecolor{ablS}{HTML}{E4F4C6}
\definecolor{ablQ}{HTML}{FBF1BF}
\definecolor{ablP}{HTML}{FCDDBF}
\definecolor{ablR}{HTML}{E3DDF5}
\providecommand{\ablbox}[2]{{\setlength{\fboxsep}{1pt}\colorbox{#1}{#2}}}

\begin{table}[t]
  \centering
  \caption{\textbf{Ablation on the endogenous memory.}
  Success rate and token reduction (\%) when recent and older observations are stored in full or as summary tokens, all trained without the exogenous memory. Rec.: the $R{=}4$ most recent observations. Old.: the older observations. F: full visual tokens. S: summary tokens. Sel.: selected by the state-refresh score. Token reduction is the reduction of visual tokens relative to keeping every observation in full, averaged over the successful episodes.}
  \label{tab:endogenous}
  \setlength{\tabcolsep}{3pt}
  \footnotesize
  \begin{tabular}{lcccccc}
    \toprule
     & \multicolumn{2}{c}{Stored as} & \multicolumn{2}{c}{\shortstack{\vphantom{X}\\Press\\Button}} & \multicolumn{2}{c}{\shortstack{Put Back\\Block\\Relocate}} \\
    \cmidrule(lr){2-3} \cmidrule(lr){4-5} \cmidrule(l){6-7}
    Variant & Rec. & Old. & \shortstack{Succ.\\(\%) $\uparrow$} & \shortstack{Token red.\\(\%) $\uparrow$} & \shortstack{Succ.\\(\%) $\uparrow$} & \shortstack{Token red.\\(\%) $\uparrow$} \\
    \midrule
    Summary only & S & S & 77 & 96.7 & 67 & 96.7 \\
    Recent + summary & F & S & 78 & 84.9 & 71 & 81.8 \\
    Full history & F & F & 90 & 0.0 & 86 & 0.0 \\
    \rowcolor{gray!15}
    EM-WAM w/o exo. & F & Sel. & \textbf{91} & 46.9 & \textbf{86} & 32.2 \\
    \bottomrule
  \end{tabular}
\end{table}

\begin{table}[t]
  \centering
  \caption{\textbf{Ablation on the exogenous memory.}
  Success rate (\%) when one component of EM-WAM is changed.}
  \label{tab:exogenous}
  \setlength{\tabcolsep}{3pt}
  \footnotesize
  \setlength{\aboverulesep}{0.2ex}
  \setlength{\belowrulesep}{0.3ex}
  \setlength{\fboxsep}{1pt}
  \begin{tabular}{>{\raggedright\arraybackslash}p{\dimexpr0.66\linewidth-6\tabcolsep\relax}
                  >{\centering\arraybackslash}p{0.2\linewidth}
                  >{\centering\arraybackslash}p{0.14\linewidth}}
    \toprule
     & \multicolumn{2}{c}{Success (\%) $\uparrow$} \\
    \cmidrule(l){2-3}
    Variant & \shortstack{Put Back\\Block Relocate} & \shortstack{Restore\\Layout} \\
    \midrule
    EM-WAM w/o exogenous & 86 & 54 \\
    \midrule
    Write selection \ablbox{ablR}{random} & 92 & 73 \\
    Denoising steps \ablbox{ablK}{$K{=}1$} & 93 & 53 \\
    Denoising steps \ablbox{ablK}{$K{=}2$} & 96 & 71 \\
    Discrepancy measure \ablbox{ablS}{cosine} & 90 & 42 \\
    Write quantile \ablbox{ablQ}{$q{=}0.7$} & 90 & 58 \\
    \midrule
    \rowcolor{gray!15}
    \mbox{EM-WAM {\scriptsize(\ablbox{ablR}{surprise},\,\ablbox{ablK}{$K{=}4$},\,\ablbox{ablS}{RMS},\,\ablbox{ablQ}{$q{=}0.9$})}} & \textbf{98} & \textbf{81} \\
    \bottomrule
  \end{tabular}
\end{table}

\begin{figure}[!b]
  \centering
  \includegraphics[width=0.88\linewidth]{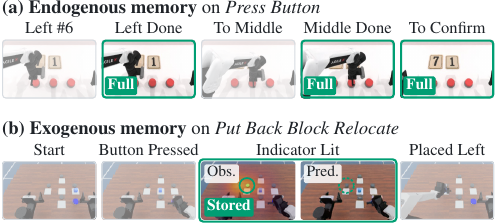}
  \caption{\textbf{What the two memories select.}
  (a)~On \textit{Press Button}, the observations right after each completed subtask are retained in full (green), and the others are compressed into summary tokens.
  (b)~On \textit{Put Back Block Relocate}, the indicator lighting up is an unexpected event that cannot be predicted from the observed history, so the exogenous memory writes this chunk into the event bank. The glow shows the regional surprise of this chunk, which peaks at the lit indicator.}
  \label{fig:memory_vis}
\end{figure}

\noindent\textbf{Endogenous Memory.}
\cref{tab:endogenous} compares EM-WAM w/o exogenous with three variants of the hybrid token cache on \textit{Press Button} (RMBench) and \textit{Put Back Block Relocate} (RMBench-Dynamic), all trained without the exogenous memory.
EM-WAM w/o exogenous keeps the $R{=}4$ most recent observations in full and, among the older ones, only those whose state-refresh score reaches $\tau_\rho$ (\cref{sec:method_endogenous}); all others are compressed into $M{=}4$ summary tokens.
\emph{Summary only} compresses every observation ($R{=}0$), \emph{Recent + summary} keeps only the $R{=}4$ most recent observations in full, and \emph{Full history} keeps every observation in full.
EM-WAM w/o exogenous performs on par with \emph{Full history} and clearly better than \emph{Recent + summary}, while reducing visual tokens by 32.2\% compared with \emph{Full history} on \textit{Put Back Block Relocate}.
The state-refresh score therefore selects the observations that need to be kept in full~(\cref{fig:memory_vis}a).

\label{sec:exp_exogenous}

\noindent\textbf{Exogenous Memory.}
\cref{tab:exogenous} ablates the components of the exogenous memory described in \cref{sec:method_exogenous} on \textit{Put Back Block Relocate} and \textit{Restore Layout}.
We change one parameter of the surprise or the memory write at a time.
\begin{itemize}[leftmargin=*,itemsep=0pt,topsep=2pt]
  \item \textit{Write selection.} At test time, the bank of the trained EM-WAM stores the same number of observations and regions, but they are taken at random from the same episode instead of being selected by the surprise. This variant still improves over EM-WAM w/o exogenous, because some random observations also contain useful information. EM-WAM clearly outperforms this variant, which shows that the regions selected by the surprise are more useful for the action expert~(\cref{fig:memory_vis}b).
  \item \textit{Denoising steps.} $K{=}1$ and $K{=}2$ instead of $4$ give a noisier prediction of the next latent, so the prediction error in this case is not a reliable indicator of surprise. The surprise no longer separates scene changes from model uncertainty, so the bank stores observations without a scene change.
  \item \textit{Discrepancy measure.} The cosine distance replaces the RMS of \cref{eq:surprise}. It compares only the directions of the two latents, so a small local change barely affects it. The RMS therefore better captures changes that cover only a few regions.
  \item \textit{Write quantile.} A lower quantile $q{=}0.7$ instead of $0.9$ writes more observations. Most of these observations contain no scene change, and they interfere with the scene changes stored in the bank. A high quantile therefore keeps the bank for the real scene changes.
\end{itemize}

%-------------------------------------------------------------------------------
\subsection{Real-World Deployment}
\label{sec:exp_real}
\begin{figure}[!htbp]
  \vspace{-4pt}
  \centering
  \includegraphics[width=\linewidth]{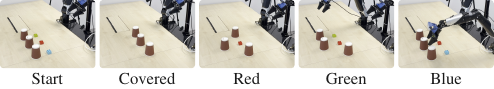}\\[2pt]
  \includegraphics[width=\linewidth]{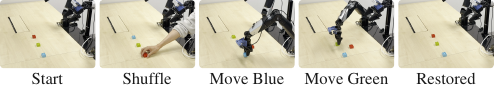}
  \caption{\textbf{Real-world deployment.} The setup on two AgileX Nero arms and example rollouts of EM-WAM on \textit{Cover Blocks} and \textit{Restore Layout}.}
  \label{fig:real_robot}
  \vspace{-6pt}
\end{figure}

We deploy EM-WAM on two AgileX Nero arms with a head camera and two wrist cameras (\cref{fig:real_robot}).
We consider two tasks, one for each memory.
In \textit{Cover Blocks}, adapted from RMBench~\cite{chen2026rmbench}, the robot covers three colored blocks with cups and then uncovers them in the order red, green, blue, so it must remember the layout it saw before the cups hid it.
In \textit{Restore Layout}, adapted from RMBench-Dynamic, a person rearranges the three blocks among their slots during the episode, and the robot must restore the original order.
For each task, we collect 300 demonstrations. All methods are trained on the same demonstrations and evaluated over the same 30 trials.

\begin{table*}[!htbp]
  \vspace{-4pt}
  \centering
  \caption[Real-world results.]{\textbf{Real-world results.} Number of trials, out of 30 per task, that end in each type of failure or in success.}
  \label{tab:real}
  \setlength{\tabcolsep}{3pt}
  \footnotesize
  \begin{tabular}{lccccccc}
    \toprule
    & \multicolumn{3}{c}{\textit{\realTaskEndo}} & \multicolumn{4}{c}{\textit{\realTaskExo}} \\
    \cmidrule(lr){2-4} \cmidrule(lr){5-8}
    Method & \shortstack{Wrong\\covering} & \shortstack{Wrong\\uncovering} & Success & \shortstack{No motion\\after shuffle} & \shortstack{Restored,\\then moved} & \shortstack{Never\\restored} & Success \\
    \midrule
    LingBot-VA~\cite{li2026lingbotva} & 1 & 13 & 16 & 3 & 8 & 8 & 11 \\
    EM-WAM w/o exogenous & 0 & 0 & \textbf{30} & 0 & 1 & 7 & 22 \\
    \rowcolor{gray!15}
    EM-WAM & 0 & 0 & \textbf{30} & 1 & 0 & 2 & \textbf{27} \\
    \bottomrule
  \end{tabular}
  \vspace{-4pt}
\end{table*}

\cref{tab:real} reports the results.
EM-WAM has the highest success on both tasks, and removing its exogenous memory lowers \textit{Restore Layout} to 22/30.
\cref{tab:real} also lists how each trial ends, and we explain each type of failure below.

\paragraph{Cover Blocks.}
A trial fails as soon as the robot covers or uncovers a cup out of order.
EM-WAM succeeds in all 30 trials, and LingBot-VA fails in 14 trials.
\begin{itemize}[leftmargin=*,itemsep=0pt,topsep=2pt]
  \item \textit{Wrong covering order.} The robot covers the blocks in a wrong order. This happens in 1 trial of LingBot-VA.
  \item \textit{Wrong uncovering order.} The robot does not uncover the blocks in the order red, green, blue. This happens in 13 trials of LingBot-VA. In 12 of them, it uncovers the red block correctly, and then uncovers the blue block instead of the green one. Its second choice follows a fixed rule based on position: it uncovers the middle cup if the red block was at either side, and the left cup otherwise. This rule is wrong for three of the six color layouts, and all these 12 failures occur in these three layouts. LingBot-VA therefore picks the cup by its position, rather than by where the green block was hidden.
\end{itemize}

\paragraph{Restore Layout.}
A trial succeeds when the robot restores the original order and then holds still for 160 actions with both grippers open.
No single action ends a trial as a failure.
A failed trial ends when the robot stays still for 160 actions or when it reaches the limit of 1{,}280 actions.
EM-WAM fails in 3 trials, and LingBot-VA fails in 19 trials.
\begin{itemize}[leftmargin=*,itemsep=0pt,topsep=2pt]
  \item \textit{No motion after the rearrangement.} The robot does not move after the person rearranges the blocks. This happens in 3 trials of LingBot-VA and 1 trial of EM-WAM.
  \item \textit{Restored, then moved out of order.} The robot restores the original order but does not stop, and moves the blocks out of order again. This happens in 8 trials of LingBot-VA and in no trial of EM-WAM.
  \item \textit{Original order never restored.} The robot keeps moving the blocks but never restores the original order. This happens in 8 trials of LingBot-VA and 2 trials of EM-WAM. In both trials of EM-WAM, it moves the blocks into another wrong order and does not correct it before the action limit.
\end{itemize}

\clearpage
%-------------------------------------------------------------------------------
\section{Conclusion}
\label{sec:conclusion}

We have presented EM-WAM, a world action model that equips its video and action experts with two complementary memories.
Guided by a state-refresh score read from the decay gates of the video expert, the endogenous memory keeps in full the informative past frames and summarizes the rest with a few tokens.
The exogenous memory writes unexpected events into an event bank, so that they keep guiding the action expert after they leave the view.
To measure how a policy maintains and updates its memory, we introduced RMBench-Dynamic, eight dual-arm tasks in which unexpected events occur during the episode.
EM-WAM sets a new state of the art on RMBench-Dynamic and on RMBench, and completes 57 of 60 trials on a real dual-arm robot.

\noindent\textbf{Limitations and Future Work.}
Currently, our EM-WAM serves as a System~1 policy and is not yet connected to a System~2 planner.
In the future, we will combine EM-WAM with a System~2 planner, design the interface between the planner and the memories of EM-WAM, and validate the combined system on long-horizon tasks.

\FloatBarrier
{
    \small
    \bibliographystyle{ieeenat_fullname}
    \bibliography{main}

\begin{thebibliography}{65}
\providecommand{\natexlab}[1]{#1}
\providecommand{\url}[1]{\texttt{#1}}
\expandafter\ifx\csname urlstyle\endcsname\relax
  \providecommand{\doi}[1]{doi: #1}\else
  \providecommand{\doi}{doi: \begingroup \urlstyle{rm}\Url}\fi

\bibitem[Aakur and Sarkar(2019)]{aakur2019perceptual}
Sathyanarayanan~N. Aakur and Sudeep Sarkar.
\newblock A perceptual prediction framework for self supervised event segmentation.
\newblock In \emph{Proceedings of the IEEE/CVF Conference on Computer Vision and Pattern Recognition}, pages 1197--1206, 2019.

\bibitem[Bai et~al.(2025)Bai, Chen, Liu, Wang, Ge, Song, Dang, Wang, Wang, Tang, Zhong, Zhu, Yang, Li, Wan, Wang, Ding, Fu, Xu, Ye, Zhang, Xie, Cheng, Zhang, Yang, Xu, and Lin]{bai2025qwen25vl}
Shuai Bai, Keqin Chen, Xuejing Liu, Jialin Wang, Wenbin Ge, Sibo Song, Kai Dang, Peng Wang, Shijie Wang, Jun Tang, Humen Zhong, Yuanzhi Zhu, Mingkun Yang, Zhaohai Li, Jianqiang Wan, Pengfei Wang, Wei Ding, Zheren Fu, Yiheng Xu, Jiabo Ye, Xi Zhang, Tianbao Xie, Zesen Cheng, Hang Zhang, Zhibo Yang, Haiyang Xu, and Junyang Lin.
\newblock {Qwen2.5-VL} technical report.
\newblock arXiv preprint arXiv:2502.13923, 2025.

\bibitem[Behrouz et~al.(2025)Behrouz, Zhong, and Mirrokni]{behrouz2025titans}
Ali Behrouz, Peilin Zhong, and Vahab Mirrokni.
\newblock {Titans}: Learning to memorize at test time.
\newblock In \emph{Advances in Neural Information Processing Systems}, pages 113506--113543, 2025.

\bibitem[Black et~al.(2025)Black, Brown, Darpinian, Dhabalia, Driess, Esmail, Equi, Finn, Fusai, Galliker, Ghosh, Groom, Hausman, Ichter, Jakubczak, Jones, Ke, LeBlanc, Levine, Li-Bell, Mothukuri, Nair, Pertsch, Ren, Shi, Smith, Springenberg, Stachowicz, Tanner, Vuong, Walke, Walling, Wang, Yu, and Zhilinsky]{black2025pi05}
Kevin Black, Noah Brown, James Darpinian, Karan Dhabalia, Danny Driess, Adnan Esmail, Michael~Robert Equi, Chelsea Finn, Niccolo Fusai, Manuel~Y. Galliker, Dibya Ghosh, Lachy Groom, Karol Hausman, Brian Ichter, Szymon Jakubczak, Tim Jones, Liyiming Ke, Devin LeBlanc, Sergey Levine, Adrian Li-Bell, Mohith Mothukuri, Suraj Nair, Karl Pertsch, Allen~Z. Ren, Lucy~Xiaoyang Shi, Laura Smith, Jost~Tobias Springenberg, Kyle Stachowicz, James Tanner, Quan Vuong, Homer Walke, Anna Walling, Haohuan Wang, Lili Yu, and Ury Zhilinsky.
\newblock $\pi_{0.5}$: a vision-language-action model with open-world generalization.
\newblock In \emph{Proceedings of The 9th Conference on Robot Learning}, pages 17--40. PMLR, 2025.

\bibitem[Cheang et~al.(2024)Cheang, Chen, Jing, Kong, Li, Li, Liu, Wu, Xu, Yang, Zhang, and Zhu]{cheang2024gr2}
Chi-Lam Cheang, Guangzeng Chen, Ya Jing, Tao Kong, Hang Li, Yifeng Li, Yuxiao Liu, Hongtao Wu, Jiafeng Xu, Yichu Yang, Hanbo Zhang, and Minzhao Zhu.
\newblock {GR-2}: A generative video-language-action model with web-scale knowledge for robot manipulation.
\newblock arXiv preprint arXiv:2410.06158, 2024.

\bibitem[Chen et~al.(2026{\natexlab{a}})Chen, Chen, Li, Tang, Su, Lu, Wan, Chen, Liu, Yan, et~al.]{chen2026robodojo}
Tianxing Chen, Yue Chen, Zixuan Li, Junyuan Tang, Kailun Su, Haoran Lu, Weijie Wan, Baijun Chen, Songling Liu, Haowen Yan, et~al.
\newblock {RoboDojo}: A unified sim-and-real benchmark for comprehensive evaluation of generalist robot manipulation policies.
\newblock arXiv preprint arXiv:2607.04434, 2026{\natexlab{a}}.

\bibitem[Chen et~al.(2026{\natexlab{b}})Chen, Chen, Chen, Cai, Liu, Li, Liang, Lin, Ge, Gu, Deng, Guo, Nian, Xie, Chen, Su, Xu, Liu, Hu, Gao, Wang, Liang, Qin, Yang, Luo, and Mu]{chen2026robotwin2}
Tianxing Chen, Zanxin Chen, Baijun Chen, Zijian Cai, Yibin Liu, Zixuan Li, Qiwei Liang, Xianliang Lin, Yiheng Ge, Zhenyu Gu, Weiliang Deng, Yubin Guo, Tian Nian, Xuanbing Xie, Qiangyu Chen, Kailun Su, Tianling Xu, Guodong Liu, Mengkang Hu, Huan-ang Gao, Kaixuan Wang, Zhixuan Liang, Yusen Qin, Xiaokang Yang, Ping Luo, and Yao Mu.
\newblock {RoboTwin} 2.0: A scalable data generator and benchmark with strong domain randomization for robust bimanual robotic manipulation.
\newblock In \emph{Proceedings of the 43rd International Conference on Machine Learning}, pages 13673--13699. PMLR, 2026{\natexlab{b}}.

\bibitem[Chen et~al.(2026{\natexlab{c}})Chen, Wang, Li, Qin, Shi, Li, Hu, Zhang, Wang, Chen, Wang, Wang, Yang, Xu, Wu, Mu, Yang, Dong, and Luo]{chen2026rmbench}
Tianxing Chen, Yuran Wang, Mingleyang Li, Yan Qin, Hao Shi, Zixuan Li, Yifan Hu, Yingsheng Zhang, Kaixuan Wang, Yue Chen, Hongcheng Wang, Junjie Wang, Tianhang Yang, Renjing Xu, Ruihai Wu, Yao Mu, Yaodong Yang, Hao Dong, and Ping Luo.
\newblock {RMBench}: Memory-dependent robotic manipulation benchmark with insights into policy design.
\newblock arXiv preprint arXiv:2603.01229, 2026{\natexlab{c}}.

\bibitem[Cherepanov et~al.(2026)Cherepanov, Kachaev, Kovalev, and Panov]{cherepanov2026mikasarobo}
Egor Cherepanov, Nikita Kachaev, Alexey~K. Kovalev, and Aleksandr~I. Panov.
\newblock Memory, benchmark \& robots: A benchmark for solving complex tasks with reinforcement learning.
\newblock In \emph{International Conference on Learning Representations}, pages 9759--9815, 2026.

\bibitem[Chung et~al.(2026)Chung, Hanyu, Nguyen, Le, Bumgarner, Nguyen, Vo, Yamazaki, Rainwater, Kieu, Nguyen, and Le]{chung2026liberomem}
Nhat Chung, Taisei Hanyu, Toan Nguyen, Huy Le, Frederick Bumgarner, Duy Minh~Ho Nguyen, Khoa Vo, Kashu Yamazaki, Chase Rainwater, Tung Kieu, Anh Nguyen, and Ngan Le.
\newblock Rethinking progression of memory state in robotic manipulation: An object-centric perspective.
\newblock In \emph{Proceedings of the AAAI Conference on Artificial Intelligence}, pages 3407--3415, 2026.

\bibitem[Dai et~al.(2026)Dai, Fu, Lee, Liu, Zhang, Yang, Finn, Fazeli, and Chai]{dai2026robomme}
Yinpei Dai, Hongze Fu, Jayjun Lee, Yuejiang Liu, Haoran Zhang, Jianing Yang, Chelsea Finn, Nima Fazeli, and Joyce Chai.
\newblock {RoboMME}: Benchmarking and understanding memory for robotic generalist policies.
\newblock In \emph{Proceedings of the 43rd International Conference on Machine Learning}, pages 22396--22450. PMLR, 2026.

\bibitem[Dao and Gu(2024)]{dao2024mamba2}
Tri Dao and Albert Gu.
\newblock Transformers are {SSMs}: Generalized models and efficient algorithms through structured state space duality.
\newblock In \emph{Proceedings of the 41st International Conference on Machine Learning}, pages 10041--10071. PMLR, 2024.

\bibitem[{Dexmal Team}(2026)]{dexmal2026dm05}
{Dexmal Team}.
\newblock {DM0.5}: An open-world foundation model for general-purpose embodied intelligence.
\newblock \url{https://www.dexmal.com/blog/dm0.5/index_en.html}, 2026.

\bibitem[Du et~al.(2023)Du, Yang, Dai, Dai, Nachum, Tenenbaum, Schuurmans, and Abbeel]{du2023unipi}
Yilun Du, Sherry Yang, Bo Dai, Hanjun Dai, Ofir Nachum, Josh Tenenbaum, Dale Schuurmans, and Pieter Abbeel.
\newblock Learning universal policies via text-guided video generation.
\newblock In \emph{Advances in Neural Information Processing Systems}, pages 9156--9172, 2023.

\bibitem[Fang et~al.(2025)Fang, Grotz, Pumacay, Wang, Fox, Krishna, and Duan]{fang2025sam2act}
Haoquan Fang, Markus Grotz, Wilbert Pumacay, Yi~Ru Wang, Dieter Fox, Ranjay Krishna, and Jiafei Duan.
\newblock {SAM2Act}: Integrating visual foundation model with a memory architecture for robotic manipulation.
\newblock In \emph{Proceedings of the 42nd International Conference on Machine Learning}, pages 15925--15942. PMLR, 2025.

\bibitem[Fei et~al.(2026)Fei, Wang, Shi, Dai, Cai, Qian, Ji, He, Zhang, Fei, Fu, Gong, and Qiu]{fei2026liberoplus}
Senyu Fei, Siyin Wang, Junhao Shi, Zihao Dai, Jikun Cai, Pengfang Qian, Li Ji, Xinzhe He, Shiduo Zhang, Zhaoye Fei, Jinlan Fu, Jingjing Gong, and Xipeng Qiu.
\newblock {LIBERO-Plus}: A progressive robustness benchmark for visual-language-action models.
\newblock In \emph{Proceedings of the IEEE/CVF Conference on Computer Vision and Pattern Recognition}, pages 38574--38583, 2026.

\bibitem[Fountas et~al.(2025)Fountas, Benfeghoul, Oomerjee, Christopoulou, Lampouras, Bou~Ammar, and Wang]{fountas2025emllm}
Zafeirios Fountas, Martin~A. Benfeghoul, Adnan Oomerjee, Fenia Christopoulou, Gerasimos Lampouras, Haitham Bou~Ammar, and Jun Wang.
\newblock Human-inspired episodic memory for infinite context {LLMs}.
\newblock In \emph{International Conference on Learning Representations}, pages 77230--77268, 2025.

\bibitem[Gu and Dao(2024)]{gu2024mamba}
Albert Gu and Tri Dao.
\newblock {Mamba}: Linear-time sequence modeling with selective state spaces.
\newblock In \emph{Conference on Language Modeling}, 2024.

\bibitem[Guo et~al.(2024)Guo, Wang, Zha, and Chen]{guo2023doremi}
Yanjiang Guo, Yen-Jen Wang, Lihan Zha, and Jianyu Chen.
\newblock {DoReMi}: Grounding language model by detecting and recovering from plan-execution misalignment.
\newblock In \emph{IEEE/RSJ International Conference on Intelligent Robots and Systems}, pages 12124--12131, 2024.

\bibitem[Han et~al.(2025)Han, Qiu, Liao, Huang, Gao, Yan, and Liu]{han2025robocerebra}
Songhao Han, Boxiang Qiu, Yue Liao, Siyuan Huang, Chen Gao, Shuicheng Yan, and Si Liu.
\newblock {RoboCerebra}: A large-scale benchmark for long-horizon robotic manipulation evaluation.
\newblock In \emph{Advances in Neural Information Processing Systems, Datasets and Benchmarks Track}, 2025.

\bibitem[Hu et~al.(2025)Hu, Guo, Wang, Chen, Wang, Zhang, Sreenath, Lu, and Chen]{hu2025vpp}
Yucheng Hu, Yanjiang Guo, Pengchao Wang, Xiaoyu Chen, Yen-Jen Wang, Jianke Zhang, Koushil Sreenath, Chaochao Lu, and Jianyu Chen.
\newblock Video prediction policy: A generalist robot policy with predictive visual representations.
\newblock In \emph{Proceedings of the 42nd International Conference on Machine Learning}, pages 24328--24346. PMLR, 2025.

\bibitem[Huang et~al.(2023)Huang, Xia, Xiao, Chan, Liang, Florence, Zeng, Tompson, Mordatch, Chebotar, Sermanet, Jackson, Brown, Luu, Levine, Hausman, and Ichter]{huang2023innermonologue}
Wenlong Huang, Fei Xia, Ted Xiao, Harris Chan, Jacky Liang, Pete Florence, Andy Zeng, Jonathan Tompson, Igor Mordatch, Yevgen Chebotar, Pierre Sermanet, Tomas Jackson, Noah Brown, Linda Luu, Sergey Levine, Karol Hausman, and Brian Ichter.
\newblock Inner monologue: Embodied reasoning through planning with language models.
\newblock In \emph{Conference on Robot Learning}, pages 1769--1782, 2023.

\bibitem[James et~al.(2020)James, Ma, Rovick~Arrojo, and Davison]{james2020rlbench}
Stephen James, Zicong Ma, David Rovick~Arrojo, and Andrew~J. Davison.
\newblock {RLBench}: The robot learning benchmark \& learning environment.
\newblock \emph{IEEE Robotics and Automation Letters}, 5\penalty0 (2):\penalty0 3019--3026, 2020.

\bibitem[{Kairos Team} et~al.(2026){Kairos Team}, Wang, You, Zhang, Huang, Fu, Zheng, Xi, Lv, Wu, Liu, Wan, Li, Yang, Li, Wang, Zhu, Zhang, Fu, Zhang, Wu, Fan, Tao, and Wang]{kairos2026kairos}
{Kairos Team}, Fei Wang, Shan You, Qiming Zhang, Tao Huang, Zuoyi Fu, Zhisheng Zheng, Yunlong Xi, Feng Lv, Xiaoming Wu, Zeyu Liu, Cong Wan, Pu Li, Ruiqing Yang, Xiaoou Li, Wei Wang, Kangkang Zhu, Yuwei Zhang, Shi Fu, Zheng Zhang, Xiaoning Wu, Xuzeng Fan, Dacheng Tao, and Xiaogang Wang.
\newblock {Kairos}: A regret-aware native world-action model stack for physical {AI}.
\newblock arXiv preprint arXiv:2606.16533, 2026.

\bibitem[Li et~al.(2026{\natexlab{a}})Li, Shen, Chen, Yang, Wang, Shi, Bing, Liu, and Knoll]{li2026rememvla}
Hang Li, Fengyi Shen, Dong Chen, Liudi Yang, Xudong Wang, Jinkui Shi, Zhenshan Bing, Ziyuan Liu, and Alois Knoll.
\newblock {ReMem-VLA: Empowering Vision-Language-Action Model with Memory via Dual-Level Recurrent Queries}.
\newblock arXiv preprint arXiv:2603.12942, 2026{\natexlab{a}}.

\bibitem[Li et~al.(2026{\natexlab{b}})Li, Zhang, Luo, Yang, Wang, Zhang, Yu, Gao, Xue, Zhou, Zhu, Ding, Shen, and Xu]{li2026lingbotva}
Lin Li, Qihang Zhang, Yiming Luo, Shuai Yang, Ruilin Wang, Luyao Zhang, Mingrui Yu, Zelin Gao, Nan Xue, Boyu Zhou, Xing Zhu, Mingyu Ding, Yujun Shen, and Yinghao Xu.
\newblock Causal world modeling for robot control.
\newblock In \emph{Robotics: Science and Systems XXII}, 2026{\natexlab{b}}.

\bibitem[Li et~al.(2025)Li, Gao, Sadigh, and Song]{li2025uva}
Shuang Li, Yihuai Gao, Dorsa Sadigh, and Shuran Song.
\newblock Unified video action model.
\newblock In \emph{Robotics: Science and Systems XXI}, 2025.

\bibitem[Lipman et~al.(2023)Lipman, Chen, Ben-Hamu, Nickel, and Le]{lipman2023flowmatching}
Yaron Lipman, Ricky T.~Q. Chen, Heli Ben-Hamu, Maximilian Nickel, and Matt Le.
\newblock Flow matching for generative modeling.
\newblock In \emph{International Conference on Learning Representations}, 2023.

\bibitem[Liu et~al.(2023{\natexlab{a}})Liu, Zhu, Gao, Feng, Liu, Zhu, and Stone]{liu2023libero}
Bo Liu, Yifeng Zhu, Chongkai Gao, Yihao Feng, Qiang Liu, Yuke Zhu, and Peter Stone.
\newblock {LIBERO}: Benchmarking knowledge transfer for lifelong robot learning.
\newblock In \emph{Advances in Neural Information Processing Systems}, pages 44776--44791, 2023{\natexlab{a}}.

\bibitem[Liu et~al.(2023{\natexlab{b}})Liu, Gong, and Liu]{liu2023rectifiedflow}
Xingchao Liu, Chengyue Gong, and Qiang Liu.
\newblock Flow straight and fast: Learning to generate and transfer data with rectified flow.
\newblock In \emph{International Conference on Learning Representations}, 2023{\natexlab{b}}.

\bibitem[Liu et~al.(2023{\natexlab{c}})Liu, Bahety, and Song]{liu2023reflect}
Zeyi Liu, Arpit Bahety, and Shuran Song.
\newblock {REFLECT}: Summarizing robot experiences for failure explanation and correction.
\newblock In \emph{Conference on Robot Learning}, pages 3468--3484, 2023{\natexlab{c}}.

\bibitem[Mees et~al.(2022)Mees, Hermann, Rosete-Beas, and Burgard]{mees2022calvin}
Oier Mees, Lukas Hermann, Erick Rosete-Beas, and Wolfram Burgard.
\newblock {CALVIN}: A benchmark for language-conditioned policy learning for long-horizon robot manipulation tasks.
\newblock \emph{IEEE Robotics and Automation Letters}, 7\penalty0 (3):\penalty0 7327--7334, 2022.

\bibitem[Miao et~al.(2026)Miao, Yu, Liu, Yang, Chen, Zheng, Wang, and Zhang]{miao2026rmwam}
Cui Miao, Junhao Yu, Ping Liu, Ning Yang, Xuhua Chen, He Zheng, Xiaodong Wang, and Tao Zhang.
\newblock {RM-WAM}: Reliable memory-augmented world action models for long-horizon robot manipulation.
\newblock arXiv preprint arXiv:2610.15305, 2026.

\bibitem[Peebles and Xie(2023)]{peebles2023dit}
William Peebles and Saining Xie.
\newblock Scalable diffusion models with transformers.
\newblock In \emph{Proceedings of the IEEE/CVF International Conference on Computer Vision}, pages 4195--4205, 2023.

\bibitem[Pham and Pham(2018)]{pham2018toppra}
Hung Pham and Quang-Cuong Pham.
\newblock A new approach to time-optimal path parameterization based on reachability analysis.
\newblock \emph{IEEE Transactions on Robotics}, 34\penalty0 (3):\penalty0 645--659, 2018.

\bibitem[Rae et~al.(2020)Rae, Potapenko, Jayakumar, Hillier, and Lillicrap]{rae2020compressive}
Jack~W. Rae, Anna Potapenko, Siddhant~M. Jayakumar, Chloe Hillier, and Timothy~P. Lillicrap.
\newblock Compressive transformers for long-range sequence modelling.
\newblock In \emph{International Conference on Learning Representations}, 2020.

\bibitem[Schlag et~al.(2021)Schlag, Irie, and Schmidhuber]{schlag2021linear}
Imanol Schlag, Kazuki Irie, and J{\"u}rgen Schmidhuber.
\newblock Linear transformers are secretly fast weight programmers.
\newblock In \emph{Proceedings of the 38th International Conference on Machine Learning}, pages 9355--9366. PMLR, 2021.

\bibitem[{SenseNova} et~al.(2026){SenseNova}, {Joint Research Team}, Cai, Cao, Chen, Chua, Dai, Deng, Deng, Diao, et~al.]{sensenova2026gpt6astra}
{SenseNova}, {Joint Research Team}, Zhongang Cai, Ziang Cao, Tailai Chen, Han~Chong Chua, Linjun Dai, Hanming Deng, Hokin Deng, Haiwen Diao, et~al.
\newblock Has {GPT-6-Astra} achieved embodied intelligence?
\newblock arXiv preprint arXiv:2610.22471, 2026.

\bibitem[Senthilkumaran et~al.(2026)Senthilkumaran, Wickramarachchi, Oltramari, and Francis]{senthilkumaran2026remm}
Revanth~Krishna Senthilkumaran, Ruwan Wickramarachchi, Alessandro Oltramari, and Jonathan Francis.
\newblock {REMM}: Revisable embodied memory for manipulation.
\newblock arXiv preprint arXiv:2610.12710, 2026.

\bibitem[Shi et~al.(2026)Shi, Xie, Liu, Sun, Liu, Wang, Zhou, Fan, Zhang, and Huang]{shi2026memoryvla}
Hao Shi, Bin Xie, Yingfei Liu, Lin Sun, Fengrong Liu, Tiancai Wang, Erjin Zhou, Haoqiang Fan, Xiangyu Zhang, and Gao Huang.
\newblock {MemoryVLA: Perceptual-Cognitive Memory in Vision-Language-Action Models for Robotic Manipulation}.
\newblock In \emph{International Conference on Learning Representations}, pages 18567--18602, 2026.

\bibitem[Shi et~al.(2025)Shi, Ichter, Equi, Ke, Pertsch, Vuong, Tanner, Walling, Wang, Fusai, Li-Bell, Driess, Groom, Levine, and Finn]{shi2025hirobot}
Lucy~Xiaoyang Shi, Brian Ichter, Michael~Robert Equi, Liyiming Ke, Karl Pertsch, Quan Vuong, James Tanner, Anna Walling, Haohuan Wang, Niccolo Fusai, Adrian Li-Bell, Danny Driess, Lachy Groom, Sergey Levine, and Chelsea Finn.
\newblock Hi robot: Open-ended instruction following with hierarchical vision-language-action models.
\newblock In \emph{International Conference on Machine Learning}, pages 54919--54933, 2025.

\bibitem[Sridhar et~al.(2026)Sridhar, Pan, Sharma, and Finn]{sridhar2026memer}
Ajay Sridhar, Jennifer Pan, Satvik Sharma, and Chelsea Finn.
\newblock {Scaling up Memory for Robotic Control via Experience Retrieval}.
\newblock In \emph{International Conference on Learning Representations}, pages 97142--97166, 2026.

\bibitem[Sun et~al.(2026{\natexlab{a}})Sun, Wang, Chen, Fang, Su, Yi, Yi, Chen, and Zhao]{sun2026memobench}
Haiyang Sun, Haoxiao Wang, Junming Chen, Weicheng Fang, Zihao Su, Jingkun Yi, Wenyou Yi, Hao Chen, and Zhou Zhao.
\newblock {MEMOBench}: A process level memory benchmark for robotic manipulation.
\newblock In \emph{Findings of the Association for Computational Linguistics: EMNLP 2026}, pages 7025--7040, 2026{\natexlab{a}}.

\bibitem[Sun et~al.(2026{\natexlab{b}})Sun, Zhang, Cao, Sun, Chen, Xu, Chen, Chen, Yang, Zhu, Hong, Xu, Pang, Yuan, and Chen]{sun2026himemwam}
Xiaoquan Sun, Ruijian Zhang, Chen Cao, Yihan Sun, Jiahui Chen, Zetian Xu, Bo Chen, Haijier Chen, Zhen Yang, Jiarun Zhu, Yijun Hong, JingZhe Xu, Jingrui Pang, Mingqi Yuan, and Jiayu Chen.
\newblock {HiMem-WAM: Hierarchical Memory-Gated World Action Models for Robotic Manipulation}.
\newblock arXiv preprint arXiv:2606.10363, 2026{\natexlab{b}}.

\bibitem[{Team Wan} et~al.(2025){Team Wan}, Wang, Ai, Wen, Mao, Xie, Chen, Yu, Zhao, Yang, Zeng, Wang, Zhang, Zhou, Wang, Chen, Zhu, Zhao, Yan, Huang, Feng, Zhang, Li, Wu, Chu, Feng, Zhang, Sun, Fang, Wang, Gui, Weng, Shen, Lin, Wang, Wang, Zhou, Wang, Shen, Yu, Shi, Huang, Xu, Kou, Lv, Li, Liu, Wang, Zhang, Huang, Li, Wu, Liu, Pan, Zheng, Hong, Shi, Feng, Jiang, Han, Wu, and Liu]{wan2025wan}
{Team Wan}, Ang Wang, Baole Ai, Bin Wen, Chaojie Mao, Chen-Wei Xie, Di Chen, Feiwu Yu, Haiming Zhao, Jianxiao Yang, Jianyuan Zeng, Jiayu Wang, Jingfeng Zhang, Jingren Zhou, Jinkai Wang, Jixuan Chen, Kai Zhu, Kang Zhao, Keyu Yan, Lianghua Huang, Mengyang Feng, Ningyi Zhang, Pandeng Li, Pingyu Wu, Ruihang Chu, Ruili Feng, Shiwei Zhang, Siyang Sun, Tao Fang, Tianxing Wang, Tianyi Gui, Tingyu Weng, Tong Shen, Wei Lin, Wei Wang, Wei Wang, Wenmeng Zhou, Wente Wang, Wenting Shen, Wenyuan Yu, Xianzhong Shi, Xiaoming Huang, Xin Xu, Yan Kou, Yangyu Lv, Yifei Li, Yijing Liu, Yiming Wang, Yingya Zhang, Yitong Huang, Yong Li, You Wu, Yu Liu, Yulin Pan, Yun Zheng, Yuntao Hong, Yupeng Shi, Yutong Feng, Zeyinzi Jiang, Zhen Han, Zhi-Fan Wu, and Ziyu Liu.
\newblock {Wan}: Open and advanced large-scale video generative models.
\newblock arXiv preprint arXiv:2503.20314, 2025.

\bibitem[Tian et~al.(2025)Tian, Yang, Zeng, Wang, Lin, Dong, and Pang]{tian2025seer}
Yang Tian, Sizhe Yang, Jia Zeng, Ping Wang, Dahua Lin, Hao Dong, and Jiangmiao Pang.
\newblock Predictive inverse dynamics models are scalable learners for robotic manipulation.
\newblock In \emph{International Conference on Learning Representations}, pages 92033--92052, 2025.

\bibitem[Torne et~al.(2026)Torne, Pertsch, Walke, Vedder, Nair, Ichter, Ren, Wang, Tang, Stachowicz, Dhabalia, Equi, Vuong, Springenberg, Levine, Finn, and Driess]{torne2026mem}
Marcel Torne, Karl Pertsch, Homer Walke, Kyle Vedder, Suraj Nair, Brian Ichter, Allen~Z. Ren, Haohuan Wang, Jiaming Tang, Kyle Stachowicz, Karan Dhabalia, Michael~Robert Equi, Quan Vuong, Jost~Tobias Springenberg, Sergey Levine, Chelsea Finn, and Danny Driess.
\newblock {MEM: Multi-Scale Embodied Memory for Vision Language Action Models}.
\newblock In \emph{Conference on Robot Learning}, 2026.

\bibitem[Villasevil et~al.(2025)Villasevil, Tang, Liu, and Finn]{torne2025pasttoken}
Marcel~Torne Villasevil, Andy Tang, Yuejiang Liu, and Chelsea Finn.
\newblock Learning long-context diffusion policies via past-token prediction.
\newblock In \emph{Proceedings of The 9th Conference on Robot Learning}, pages 1744--1755. PMLR, 2025.

\bibitem[Wang et~al.(2026)Wang, Zhang, Chen, Lin, Wang, and Qi]{wang2026whentotrust}
Rui Wang, Yue Zhang, Canyang Chen, Jiehong Lin, Zhongrui Wang, and Xiaojuan Qi.
\newblock When to trust imagination: Adaptive action execution for world action models.
\newblock arXiv preprint arXiv:2605.06222, 2026.

\bibitem[Wen et~al.(2020)Wen, Lin, Darrell, Jayaraman, and Gao]{wen2020copycat}
Chuan Wen, Jierui Lin, Trevor Darrell, Dinesh Jayaraman, and Yang Gao.
\newblock Fighting copycat agents in behavioral cloning from observation histories.
\newblock In \emph{Advances in Neural Information Processing Systems}, pages 2564--2575, 2020.

\bibitem[Wen et~al.(2021)Wen, Lin, Qian, Gao, and Jayaraman]{wen2021keyframe}
Chuan Wen, Jierui Lin, Jianing Qian, Yang Gao, and Dinesh Jayaraman.
\newblock Keyframe-focused visual imitation learning.
\newblock In \emph{Proceedings of the 38th International Conference on Machine Learning}, pages 11123--11133. PMLR, 2021.

\bibitem[Wu et~al.(2024)Wu, Jing, Cheang, Chen, Xu, Li, Liu, Li, and Kong]{wu2024gr1}
Hongtao Wu, Ya Jing, Chilam Cheang, Guangzeng Chen, Jiafeng Xu, Xinghang Li, Minghuan Liu, Hang Li, and Tao Kong.
\newblock Unleashing large-scale video generative pre-training for visual robot manipulation.
\newblock In \emph{International Conference on Learning Representations}, pages 10641--10662, 2024.

\bibitem[Xiao et~al.(2024)Xiao, Tian, Chen, Han, and Lewis]{xiao2024streamingllm}
Guangxuan Xiao, Yuandong Tian, Beidi Chen, Song Han, and Mike Lewis.
\newblock Efficient streaming language models with attention sinks.
\newblock In \emph{International Conference on Learning Representations}, pages 21875--21895, 2024.

\bibitem[Xue et~al.(2026)Xue, Zhang, Ma, Yao, Cui, and Wang]{xue2026hitacwam}
Chao Xue, Chaofan Zhang, Wenxuan Ma, Guocai Yao, Shaowei Cui, and Shuo Wang.
\newblock {HiTac-WAM}: A hierarchical tactile world action model for contact-rich robot manipulation.
\newblock arXiv preprint arXiv:2608.19574, 2026.

\bibitem[Yang et~al.(2026{\natexlab{a}})Yang, Tu, Yang, Mao, Dong, Chen, Peng, Xiong, Cao, Dai, Zhou, Mu, and Wang]{yang2026eventvla}
Ganlin Yang, Zhangzheng Tu, Yuqiang Yang, Sitong Mao, Junyi Dong, Tianxing Chen, Jiaqi Peng, Jing Xiong, Jiafei Cao, Jifeng Dai, Wengang Zhou, Yao Mu, and Tai Wang.
\newblock {EventVLA}: Event-driven visual evidence memory for long-horizon vision-language-action policies.
\newblock In \emph{Conference on Robot Learning}, 2026{\natexlab{a}}.

\bibitem[Yang et~al.(2024)Yang, Wang, Zhang, Shen, and Kim]{yang2024deltanet}
Songlin Yang, Bailin Wang, Yu Zhang, Yikang Shen, and Yoon Kim.
\newblock Parallelizing linear transformers with the delta rule over sequence length.
\newblock In \emph{Advances in Neural Information Processing Systems}, pages 115491--115522, 2024.

\bibitem[Yang et~al.(2025)Yang, Kautz, and Hatamizadeh]{yang2025gateddeltanet}
Songlin Yang, Jan Kautz, and Ali Hatamizadeh.
\newblock Gated delta networks: Improving {Mamba2} with delta rule.
\newblock In \emph{International Conference on Learning Representations}, pages 29687--29707, 2025.

\bibitem[Yang et~al.(2026{\natexlab{b}})Yang, Mu, Wei, Lu, Li, Xu, Xue, Yuan, Lin, Pang, and Xu]{yang2026memorywam}
Sizhe Yang, Juncheng Mu, Tianming Wei, Chenhao Lu, Xiaofan Li, Linning Xu, Zhengrong Xue, Zhecheng Yuan, Dahua Lin, Jiangmiao Pang, and Huazhe Xu.
\newblock {MemoryWAM: Efficient World Action Modeling with Persistent Memory}.
\newblock arXiv preprint arXiv:2606.20562, 2026{\natexlab{b}}.

\bibitem[Yu et~al.(2020)Yu, Quillen, He, Julian, Hausman, Finn, and Levine]{yu2020metaworld}
Tianhe Yu, Deirdre Quillen, Zhanpeng He, Ryan Julian, Karol Hausman, Chelsea Finn, and Sergey Levine.
\newblock {Meta-World}: A benchmark and evaluation for multi-task and meta reinforcement learning.
\newblock In \emph{Conference on Robot Learning}, pages 1094--1100, 2020.

\bibitem[Yuan et~al.(2026)Yuan, Dong, Liu, and Zhao]{yuan2026fastwam}
Tianyuan Yuan, Zibin Dong, Yicheng Liu, and Hang Zhao.
\newblock {Fast-WAM}: Do world action models need test-time future imagination?
\newblock In \emph{Advances in Neural Information Processing Systems}, 2026.

\bibitem[Zhang et~al.(2026)Zhang, Ito, Hoshino, Ikehata, and Sato]{zhang2026wcd}
Chuhan Zhang, Seiji Ito, Kenta Hoshino, Satoshi Ikehata, and Ikuro Sato.
\newblock World-coherent decoding: Self-verifying test-time planning for world action models.
\newblock arXiv preprint arXiv:2609.02159, 2026.

\bibitem[Zhang et~al.(2023)Zhang, Sheng, Zhou, Chen, Zheng, Cai, Song, Tian, R{\'e}, Barrett, Wang, and Chen]{zhang2023h2o}
Zhenyu Zhang, Ying Sheng, Tianyi Zhou, Tianlong Chen, Lianmin Zheng, Ruisi Cai, Zhao Song, Yuandong Tian, Christopher R{\'e}, Clark Barrett, Zhangyang Wang, and Beidi Chen.
\newblock {H2O}: Heavy-hitter oracle for efficient generative inference of large language models.
\newblock In \emph{Advances in Neural Information Processing Systems}, pages 34661--34710, 2023.

\bibitem[Zhao et~al.(2026)Zhao, Shan, Yang, Lei, Zhao, Fan, Feng, and Jia]{zhao2026dram}
Xinyu Zhao, Yixiang Shan, Tao Yang, Runyu Lei, Yiming Zhao, Jiaxin Fan, Zongbao Feng, and Peng Jia.
\newblock {DRAM}: Delta-rule recurrent associative memory for robot manipulation policies.
\newblock arXiv preprint arXiv:2609.32453, 2026.

\bibitem[Zhou et~al.(2025)Zhou, Su, Chi, Zhang, Wang, Huang, Sheng, and Wang]{zhou2025codeasmonitor}
Enshen Zhou, Qi Su, Cheng Chi, Zhizheng Zhang, Zhongyuan Wang, Tiejun Huang, Lu Sheng, and He Wang.
\newblock {Code-as-Monitor}: Constraint-aware visual programming for reactive and proactive robotic failure detection.
\newblock In \emph{Proceedings of the IEEE/CVF Conference on Computer Vision and Pattern Recognition}, pages 6919--6929, 2025.

\bibitem[Zhu et~al.(2025)Zhu, Yu, Feng, Burchfiel, Shah, and Gupta]{zhu2025uwm}
Chuning Zhu, Raymond Yu, Siyuan Feng, Benjamin Burchfiel, Paarth Shah, and Abhishek Gupta.
\newblock Unified world models: Coupling video and action diffusion for pretraining on large robotic datasets.
\newblock In \emph{Robotics: Science and Systems XXI}, 2025.

\end{thebibliography}
}

\FloatBarrier
\clearpage
\setcounter{arxivsupppage}{\value{page}}
\clearpage
\setcounter{page}{1}
\maketitlesupplementary
\appendix
\setcounter{figure}{0}
\setcounter{table}{0}
\renewcommand{\thefigure}{\Alph{figure}}
\renewcommand{\thetable}{\Alph{table}}
\renewcommand{\theHfigure}{supp.\Alph{figure}}
\renewcommand{\theHtable}{supp.\Alph{table}}

This supplementary material is organized as follows. \cref{sec:supp_rmbenchpp} describes RMBench-Dynamic, including its tasks, demonstrations, evaluation protocol and execution interface. \cref{sec:supp_rmbench} reports the per-task results on RMBench and the settings of the compared methods and \cref{sec:supp_hyperparams} gives the implementation details of EM-WAM.

\section{\texorpdfstring{RMBench-Dynamic}{RMBench-Dynamic} Details}
\label{sec:supp_rmbenchpp}
\cref{tab:supp_rmbenchpp_overview} lists the language instruction, the per-seed variation and the demonstration statistics of the eight tasks of RMBench-Dynamic. Below, we describe the execution interface (\cref{sec:supp_rmbenchpp_exec}), the demonstrations and the evaluation protocol (\cref{sec:supp_rmbenchpp_protocol}), and the rules of each task (\cref{sec:supp_rmbenchpp_tasks}).
\begin{table*}[!htbp]
  \centering
  \caption{\textbf{RMBench-Dynamic tasks}, grouped by primary event type.
  Instruction: language instruction of the task.
  Varied per seed: what changes across seeds.
  Frames: average number of frames of a demonstration, recorded at 25\,Hz with 50 demonstrations per task.
  Length: mean, minimum and maximum length of a demonstration.}
  \label{tab:supp_rmbenchpp_overview}
  \label{tab:rmbenchpp_tasks}
  \label{tab:supp_rmbenchpp_stats}
  \label{tab:supp_rmbenchpp_instr}
  \label{tab:supp_rmbenchpp_seeds}
  \footnotesize
  \setlength{\tabcolsep}{2.5pt}
  \begin{tabular}{@{}>{\raggedright\arraybackslash}p{0.15\linewidth} >{\raggedright\arraybackslash}p{0.37\linewidth} >{\raggedright\arraybackslash}p{0.24\linewidth} >{\raggedleft\arraybackslash}p{0.055\linewidth} >{\raggedleft\arraybackslash}p{0.04\linewidth} >{\raggedleft\arraybackslash}p{0.04\linewidth} >{\raggedleft\arraybackslash}p{0.04\linewidth}@{}}
    \toprule
    & & & & \multicolumn{3}{c}{Length (s)} \\
    \cmidrule(l){5-7}
    Task & Instruction & Varied per seed & Frames & Mean & Min & Max \\
    \midrule
    \multicolumn{7}{@{}l}{\textit{Goal revision}} \\
    Press Button Update & Press each button with the left arm until its registered presses match its number card. Then press the confirm button once with the right arm. & Targets (1--9 per button); press, button and size (1--3) of the increase & 514.3 & 20.6 & 14.0 & 25.0 \\
    Put Back Block Relocate & Move the block from its mat to the center and press the button. Place the block on the mat whose indicator lights up, or return it to its original mat if no indicator lights up. & Original mat; indicated mat, if any & 414.1 & 16.6 & 15.3 & 17.7 \\
    Rearrange Blocks Interrupt & Move block A to the empty mat and press the button. If the signal is green, move block B to the center; if it is yellow, move block A back to the center. & Mat of B; signal color & 395.7 & 15.8 & 15.2 & 16.4 \\
    Switch Panel Revision & Press the target button to see the requested pattern, set the switches to match it, and press the target button again to check. Follow the latest target and repair any switch that springs back. & Initial switch states; targets and whether a correct check revises the target; switches that spring back and their delays & 656.5 & 26.3 & 9.7 & 42.0 \\
    \midrule
    \multicolumn{7}{@{}l}{\textit{Ineffective action}} \\
    Press Button No-op & Press each button with the left arm until its registered presses match its number card; a press is registered only if the indicator lights up. Then press the confirm button once with the right arm. & Targets (1--9 per button); number (1--3) and positions of the unregistered presses & 545.7 & 21.8 & 12.3 & 28.6 \\
    \midrule
    \multicolumn{7}{@{}l}{\textit{Perturbation}} \\
    Restore Layout & Remember the colors of the three blocks, then press and release the button. Rearrange the blocks to restore the initial color layout, using the yellow buffer area if needed. & Initial color order; swapped pair & 737.1 & 29.5 & 22.2 & 41.8 \\
    Cover Blocks Shuffle & Cover the three blocks with the cups. After the covered blocks are exchanged, uncover the red, green and blue blocks in this order. & Initial block order & 1{,}027.9 & 41.1 & 40.6 & 41.7 \\
    \midrule
    \multicolumn{7}{@{}l}{\textit{Deferred feedback}} \\
    Ordered Card Memory & Press the buttons to open the four cards and find their values. Then open the cards in the order 1, 2, 3, 4. & Assignment of the values 1--4 to the cards & 523.4 & 20.9 & 12.4 & 26.2 \\
    \midrule
    All tasks & & & 601.8 & 24.1 & 9.7 & 42.0 \\
    \bottomrule
  \end{tabular}
\end{table*}
\FloatBarrier

\subsection{Platform and Execution}
\label{sec:supp_rmbenchpp_exec}

\paragraph{Observations and actions.}
RMBench-Dynamic is built in Isaac Sim with two ARX X5 arms.
The policy observes three RGB cameras (head, left wrist, and right wrist) at $480 \times 640$ and outputs a 14-dimensional absolute joint target, \ie, six joint positions and one gripper command for each arm.
Each task has one language instruction (\cref{tab:supp_rmbenchpp_instr}).

\paragraph{Fixed-rate executor.}
The physics simulation runs at 250\,Hz.
Each action is held for ten physics steps (40\,ms), after which a new observation is rendered; rendering does not advance the simulation.
The scripted expert and the evaluated policy use the same executor.
To verify the interface, we re-collected one seed per task and replayed the recorded expert actions independently: the action and observation ticks, the order and ticks of the task events, and the recorded states matched exactly in all eight tasks.

\paragraph{Execution interface of RMBench.}
RMBench inherits the action interface of RoboTwin~2.0~\cite{chen2026robotwin2}.
Its demonstrations store one frame every 15 physics steps of the 250\,Hz simulation.
At evaluation, the default executor converts each predicted joint target into a separate trajectory from the current state, time-parameterized with TOPP-RA~\cite{pham2018toppra} at the physics rate, and executes it before the next observation.
The number of physics steps per action therefore varies with the planned trajectory, whereas the demonstrations use a fixed interval.
As a result, a brief cue can span a different number of observations in the demonstrations and at evaluation.

\subsection{\texorpdfstring{Demonstrations and Evaluation}{Demonstrations and Evaluation}}
\label{sec:supp_rmbenchpp_protocol}

\paragraph{Scripted expert.}
Demonstrations are generated by a scripted expert that plans motions with privileged simulator state.
It never reads the per-seed event schedule: it reads each cue only while it is shown, \eg, an indicator or signal while the button is pressed or a card value while the card is open, and it reacts to each event only after it occurs.
It does use privileged state that a learned policy has to remember, such as the registered press counts and the block under each cup.
It succeeds on all 50 training seeds (1--50) of every task, and each demonstration ends at the first action at which the task is complete.

\paragraph{Evaluation.}
Each task is evaluated on 100 held-out seeds (100001--100100), whose event schedules are fixed in advance. An episode ends when the task succeeds, when the policy breaks a task rule (\cref{sec:supp_rmbenchpp_tasks}), or after 3{,}000 actions (120\,s). We report the success rate over the 100 seeds.

\subsection{Tasks}
\label{sec:supp_rmbenchpp_tasks}

The tasks follow the order of \cref{tab:supp_rmbenchpp_overview}. A block counts as placed when it rests on the table within 2.8\,cm of the center of its target position and neither arm touches it.

\begin{itemize}[leftmargin=*,itemsep=2pt]
  \item \textbf{Press Button Update.} Extends Press Button of RMBench~\cite{chen2026rmbench}. Number cards give the required number of registered presses for each of two buttons, which are pressed with the left arm; an indicator acknowledges every registered press. On one randomly selected press, one target increases immediately by one to three, and the presses already registered remain counted. The task succeeds when the right arm presses the confirm button once, with both counts matching the current targets. The episode fails if the policy confirms with wrong counts or presses a counting button with the right arm.
  \item \textbf{Put Back Block Relocate.} Extends Put Back Block of RMBench. The policy moves the block from its original mat to the center and presses a button; while the button is pressed, a green indicator beside one mat may light up. The block must be placed on the indicated mat, or returned to its original mat if no indicator lights up. The episode fails if the button is pressed before the block is at the center.
  \item \textbf{Rearrange Blocks Interrupt.} Extends Rearrange Blocks of RMBench. Of two identical blocks, A starts between the two mats and B starts on one of them. The policy moves A to the empty mat and presses a button, which shows a green or yellow signal while pressed. Green requires moving B to the center; yellow requires moving A back to the center, and the other block stays on its mat. The episode fails if the button is pressed before A is on the empty mat.
  \item \textbf{Switch Panel Revision.} Pressing the target button shows the requested red/green switch pattern; releasing it shows the actual switch states. The policy sets the switches and presses the target button to check. A correct check may reveal one revised target, and up to two set switches can spring back 0.48, 0.72 or 0.96\,s after the arm releases them and must be repaired. The target can be re-queried at any time. The task succeeds when a check of the latest target is accepted with all spring-backs repaired.
  \item \textbf{Press Button No-op.} Extends Press Button of RMBench. The counting task is the same, but one to three randomly chosen presses are not registered: the count does not increase and the indicator stays off. Such presses can occur late in the sequence and consecutively. The task succeeds when the right arm confirms once, with both registered counts matching the targets. The failure rules of Press Button Update apply. This task is particularly difficult. In the other tasks, the event is a visual change, such as a lit indicator or two swapped colors. Here, the event is a change that does not happen. An unregistered press looks the same as a registered one, and only its acknowledgement is missing. The policy therefore has to notice the absence of a brief cue, and a single miscounted press on either button makes the episode fail.
  \item \textbf{Restore Layout.} The policy memorizes the initial layout of a red, a green, and a blue block. While a button is pressed, all blocks appear black; on the first release, two colors are swapped without moving the blocks. The policy must rearrange the blocks to restore the initial color layout, using a yellow buffer area if needed. The task succeeds when every color is placed back at its initial position.
  \item \textbf{Cover Blocks Shuffle.} Extends Cover Blocks of RMBench. The policy covers three colored blocks with cups. Starting 4\,s after all blocks are covered and both grippers are open, the environment exchanges the covered blocks twice, which takes 1.8\,s in total. The policy then uncovers the red, green, and blue blocks in this order. The task succeeds when all blocks are clear of the cups and both grippers are released for 0.4\,s; earlier reveals alone do not count. The episode fails if a block is uncovered out of order.
  \item \textbf{Ordered Card Memory.} Four face-down cards are opened by their buttons in any order. A correct selection stays face up; a wrong selection is shown for 0.96\,s, after which all cards close and progress restarts at 1, while the mapping stays fixed. The task succeeds when 1, 2, 3, and 4 are revealed in order.
\end{itemize}

%-------------------------------------------------------------------------------
\section{Detailed Results on RMBench}
\label{sec:supp_rmbench}

\cref{tab:supp_rmbench} lists the per-task success rates on RMBench behind the group averages in Tab.~2 of the main paper.
The results of the other methods are taken from MemoryWAM~\cite{yang2026memorywam}, and we follow its setting for a fair comparison with LingBot-VA~\cite{li2026lingbotva} and MemoryWAM. In particular, we follow MemoryWAM on two tasks:
\begin{itemize}[leftmargin=*,itemsep=0pt,topsep=2pt]
  \item \textit{Observe \& Pickup.} We compare versions without pretraining, so EM-WAM, MemoryWAM and LingBot-VA are all trained without pretraining on this task. The success rates of MemoryWAM (5\%) and LingBot-VA (3\%) without pretraining are taken from MemoryWAM.
  \item \textit{Swap T.} As noted in MemoryWAM, LingBot-VA conditions its actions on the action history. We therefore add the action history to the action expert of EM-WAM on this task.
\end{itemize}

\begin{table}[!htbp]
  \centering
  \caption{\textbf{Per-task results on RMBench.}
  Success rate (\%) per task; Tab.~2 of the main paper reports the group averages.
  Best results are in \textbf{bold}.
  Tasks are grouped by memory complexity, $\mathrm{M}(1)$ and $\mathrm{M}(n)$, as in RMBench~\cite{chen2026rmbench}.
  The settings of the compared methods are described in \cref{sec:supp_rmbench}.}
  \label{tab:supp_rmbench}
  \setlength{\tabcolsep}{3.5pt}
  \footnotesize
  \begin{tabular}{lcccccccccc}
    \toprule
     & \multicolumn{5}{c}{$\mathrm{M}(1)$} & \multicolumn{4}{c}{$\mathrm{M}(n)$} & \\
    \cmidrule(lr){2-6} \cmidrule(lr){7-10}
    Method
      & \shortstack{Observe\\\& Pickup}
      & \shortstack{Rearrange\\Blocks}
      & \shortstack{Put Back\\Block}
      & \shortstack{Swap\\Blocks}
      & \shortstack{Swap\\T}
      & \shortstack{Battery\\Try}
      & \shortstack{Blocks\\Ranking Try}
      & \shortstack{Cover\\Blocks}
      & \shortstack{Press\\Button}
      & Avg. \\
    \midrule
    $\pi_{0.5}$~\cite{black2025pi05}       & \textbf{9} & 13 & 11 & 24 & 15 & 16 & 6 & 0 & 0 & 10.4 \\
    Fast-WAM~\cite{yuan2026fastwam} & 0 & 0 & 0 & 0 & 7 & 20 & 26 & 0 & 0 & 5.9 \\
    LingBot-VA~\cite{li2026lingbotva} & 3 & \textbf{100} & \textbf{100} & 99 & 88 & 41 & \textbf{100} & 79 & 84 & 77.1 \\
    MemoryWAM~\cite{yang2026memorywam} & 5 & \textbf{100} & \textbf{100} & \textbf{100} & \textbf{94} & 41 & \textbf{100} & 98 & 87 & 80.6 \\
    \midrule
    EM-WAM w/o exogenous      & 1 & \textbf{100} & \textbf{100} & \textbf{100} & 92 & \textbf{44} & \textbf{100} & \textbf{100} & \textbf{91} & 80.9 \\
    \rowcolor{gray!15}
    EM-WAM                             & 8 & \textbf{100} & \textbf{100} & \textbf{100} & 92 & \textbf{44} & \textbf{100} & \textbf{100} & \textbf{91} & \textbf{81.7} \\
    \bottomrule
  \end{tabular}
\end{table}
\FloatBarrier

%-------------------------------------------------------------------------------
\section{\texorpdfstring{Implementation Details}{Implementation Details}}
\label{sec:supp_hyperparams}

\cref{tab:supp_hyperparams} lists the hyperparameters of EM-WAM, grouped by component. The same values are used on RMBench and RMBench-Dynamic unless stated otherwise.

We train a separate model for each task. Both stages use AdamW with a batch of 8 episodes and compute the gradients over the full episode, with the attention of each step masked to its cache (Sec.~3.3 of the main paper). Stage~1 uses a learning rate of $2 \times 10^{-4}$. Stage~2 starts from the Stage-1 weights and uses a learning rate of $10^{-4}$. The camera images are resized without cropping: the head view to $256 \times 320$ and each wrist view to $128 \times 160$ with area interpolation, tiled into a $384 \times 320$ mosaic, which is resized to $192 \times 160$ with bicubic interpolation. The action expert has the same 32 layers as the video expert with a hidden width of 640, and it is initialized from the video expert by interpolation along the hidden dimension. In the real-world experiments, we collect 300 demonstrations per task (Sec.~5.4 of the main paper).

\begin{table}[!htbp]
  \centering
  \caption{\textbf{Hyperparameters of EM-WAM.}}
  \label{tab:supp_hyperparams}
  \footnotesize
  \begin{tabular}{@{}ll@{}}
    \toprule
    Hyperparameter & Value \\
    \midrule
    \multicolumn{2}{@{}l}{\textit{Input and architecture}} \\
    Input resolution & $192 \times 160$ \\
    Video expert & Kairos~\cite{kairos2026kairos} \\
    Action expert layers & 32 \\
    Action expert width & 640 \\
    Action horizon $h$ & 16 \\
    \midrule
    \multicolumn{2}{@{}l}{\textit{Endogenous memory}} \\
    Recent window $R$ & 4 \\
    Visual tokens per observation $L$ & 120 \\
    Summary tokens per observation $M$ & 4 \\
    State-refresh threshold $\tau_\rho$ & 1.5 \\
    \midrule
    \multicolumn{2}{@{}l}{\textit{Exogenous memory}} \\
    Memory slots $N_m$ & 8 \\
    Event tokens per observation $P$ & 8 \\
    Region size & $2 \times 2$ \\
    Regions per observation $N_r$ & 30 \\
    Write quantile $q$ & 0.9 \\
    Surprise buffer size & 8192 \\
    Denoising steps $K$ & 4 \\
    Event token width & 512 \\
    Memory read heads & 8 \\
    \bottomrule
  \end{tabular}
\end{table}

%-------------------------------------------------------------------------------

\end{document}